\documentclass[11pt]{article}

\usepackage{acl}

\usepackage{times}
\usepackage{latexsym}
\usepackage[T1]{fontenc}
\usepackage[utf8]{inputenc}
\usepackage{microtype}
\usepackage{inconsolata}
\usepackage{amsmath}
\usepackage{amssymb}
\usepackage{graphicx}
\usepackage{booktabs}
\usepackage{array}
\usepackage{multirow}
\usepackage{enumitem}
\usepackage{placeins}
\usepackage{float}

\title{Recognition–Refusal Misalignment in LLMs: Why Models Answer Structurally Unanswerable Questions}

\author{
  Yucheng Du \\
  University of Southern California \\
  \texttt{yuchengd@usc.edu}
  \And
  Xiyang (Sean) Hu\thanks{Corresponding author.} \\
  Arizona State University \\
  \texttt{xiyanghu@asu.edu}
}

\begin{document}
\maketitle

\begin{abstract}
\looseness=-1 Large language models often answer structurally unanswerable questions, such as computing $\cot(-540^\circ)$ or evaluating \texttt{(1).startswith("1")}, instead of abstaining. We ask whether this failure reflects missing recognition or failed routing from recognition to abstention. Across instruction-tuned models from 1.7B to 70B parameters, a single linear direction in the hidden state separates answerable from structurally impossible math and code prompts, showing that models represent impossibility before generation. Yet this recognition direction is nearly orthogonal to the canonical safety-refusal direction that mediates trained harmful-content refusal. An in-domain behavior-defined invalidity-aware direction is closer to recognition, but only partially aligned with it, and remains near-orthogonal to safety refusal. Generation-time steering along the recognition direction changes invalidity-aware behavior bidirectionally and dose-responsively on structural math and code cells, while random directions do not. Base/instruct comparisons further show that the low-cosine geometry is already present at the pretraining endpoint. The confident-on-impossible failure is therefore better explained as a routing failure than as an encoding failure: the model has a usable ``no admissible answer'' signal, but the safety-refusal pathway is not aligned to use it.
\end{abstract}

\section{Introduction}
\label{sec:intro}
LLMs sometimes answer questions with no valid answer. In our math/code benchmarks, $\cot(-540^\circ)$ is undefined and \texttt{(1).startswith("1")} raises an attribute error; nevertheless, clean-baseline generations can return answer-like outputs such as $\cot(-540^\circ)=0$ or \texttt{True} rather than abstain.\footnote{Code, datasets, aggregate experiment artifacts, and analysis scripts are available in the \href{https://github.com/yucheng-du/recognition-refusal-misalignment}{public project repository}.}

\begin{figure}[t]
  \centering
  \includegraphics[width=\columnwidth]{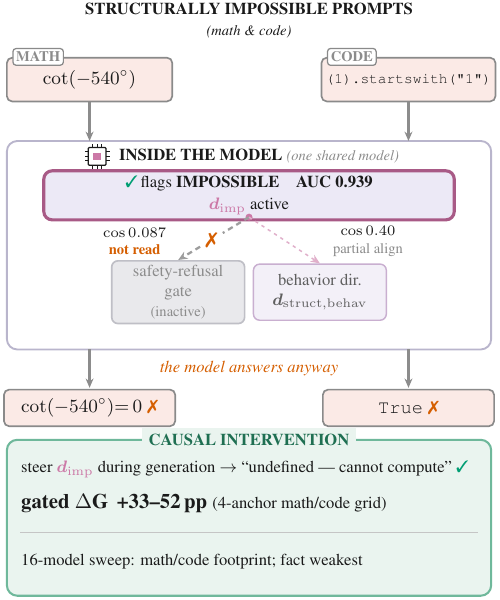}
  \caption{\textbf{A linearly accessible pre-generation signal does not guarantee action.} Structurally impossible math and code prompts activate $d_{\mathrm{imp}}$, yet the model still returns answer-like outputs (examples abbreviate real \texttt{math800}/\texttt{code800} prompts and representative clean-baseline failures). The figure summarizes the safety-refusal mismatch, the partial in-domain behavior alignment, and the causal intervention; \S\ref{sec:findings-ortho} and Fig.~\ref{fig:orthogonality} quantify the geometry.}
  \label{fig:conceptual}
\end{figure}

This behavior has two different possible explanations. The model may fail to represent the impossibility before generation, in which case abstention is unavailable. Or the model may represent the impossibility, but that signal may not be routed into the mechanism that produces abstention.

Prior work gives a concrete geometric candidate for one trained abstention mechanism: safety refusal is mediated by a single residual-stream direction $d_{\mathrm{ref,safety}}$ \citep{arditi2024refusal}. If structural-impossibility recognition reuses that existing safety-refusal pathway, an impossibility-recognition direction $d_{\mathrm{imp}}$ should align with it, i.e., $\cos(d_{\mathrm{imp}}, d_{\mathrm{ref,safety}}) \approx 1$. Because safety refusal need not be the only abstention route, we also measure an in-domain behavior-defined invalidity-aware direction in \S\ref{sec:findings-ortho}. The safety-refusal direction remains the literature-grounded comparator for trained refusal.

The evidence supports the routing account. In an 11-model main grid spanning 1.7B--70B parameters and two structural-impossibility domains (math and code; 22 model--dataset cells), a one-dimensional null-space MeanDiff probe separates answerable (A) from unanswerable (U) prompts with mean AUC \textbf{0.939}. Thus, a low-capacity reader can recover the impossibility distinction from a single residual-stream direction before generation. However, this direction is nearly orthogonal to $d_{\mathrm{ref,safety}}$, with mean cosine \textbf{0.087}. Steering along $d_{\mathrm{imp}}$ changes invalidity-aware behavior bidirectionally and dose-responsively on anchor-quality structural cells, with signal-minus-random gated flip rates of $+33$ to $+52$pp. Finally, paired base/instruct comparisons show that the low-cosine geometry largely predates instruction tuning.

Throughout the paper, ``the model knows'' is shorthand for a precise representational claim: the pre-generation hidden state contains a linearly accessible structural-impossibility signal. It does not mean that the model will use that signal in its normal generation policy. Steering along the recognition direction can change abstention behavior, but the canonical safety-refusal direction is not aligned with it. The model's trained safety-refusal pathway therefore reads a different representational axis from the one that carries structural-impossibility recognition.

The scope is structural impossibility. Math and code prompts have formally checkable rules, and matched A/U pairs can differ in a specific diagnosable feature. In this structural setting, where the analysis is cleanest, recognition's near-orthogonality to safety refusal gives a representation-level account of the kind of output-level unreasonable-math failure documented by \citet{ma2026umpr}. We use fact800 only as an epistemic-unanswerability transfer boundary and FalseQA only as a false-premise transfer boundary. These task types should not be collapsed into one generic unanswerability category: their ground truth, matched-pair construction, and intervention behavior differ.

\section{Problem Setup}
\label{sec:background}
\looseness=-1 \textbf{Three classes of unanswerability.} A question may lack an acceptable answer for distinct reasons, and those reasons determine which mechanisms are relevant. We therefore distinguish the three classes below rather than collapsing them into a single ``unanswerable'' category.

\textit{Structural impossibility} is the main setting. A prompt violates a formal rule, so no admissible answer exists: examples include $x \div 0$, $\sqrt{x}$ for $x<0$ over $\mathbb{R}$, the inverse of a singular matrix, and a Python \texttt{TypeError}. Ground truth is verifiable from the rules alone, and matched A/U pairs can differ only in a formally diagnosable feature. We use math800 (16 categories) and code800 (8 categories) for this setting. \textit{Epistemic unanswerability} means that a correct answer could exist, but the provided evidence does not determine it. We operationalize this as \textit{fact800}, paragraph-matched SQuAD~2.0 pairs whose U question's answer is absent from the shared passage. We use fact800 only as a causation and transfer contrast (\S\ref{sec:findings-causal}, \S\ref{sec:char}), not as central evidence. \textit{False-premise questions} assume a false fact, as in ``Why is CO$_2$ composed of oxygen?'' The desired behavior is to reject the premise. We use FalseQA only as a zero-shot transfer boundary (\S\ref{sec:char}), not for intervention.

The three classes have different ground-truth definitions, different pair constructions, and different behavioral signals. Accordingly, claims in \S\ref{sec:findings} should be read as claims about structural impossibility unless fact800 or FalseQA is explicitly named.

\textbf{Safety-refusal direction.} \citet{arditi2024refusal} show that safety refusal in instruction-tuned LLMs is mediated by a single residual-stream direction $d_{\mathrm{ref,safety}} = \mu_{\mathrm{harmful}} - \mu_{\mathrm{harmless}}$. We adopt the same MeanDiff construction and add behavior verification; 20 of 22 main-grid cells pass this verification (\S\ref{sec:method}).

\textbf{Research question.} Does the model encode an impossibility direction $d_{\mathrm{imp}}$, and how does that direction relate to $d_{\mathrm{ref,safety}}$? \S\ref{sec:findings} answers four sub-questions: whether $d_{\mathrm{imp}}$ exists, how it relates to safety refusal and to in-domain invalidity-aware behavior, whether the measured angle is produced by post-training, and whether $d_{\mathrm{imp}}$ is causally active on generation behavior.

\section{Method Sketch}
\label{sec:method}
\label{sec:method-cosnsrt}
\label{sec:method-hoau}
\label{sec:method-dref}
\label{sec:method-ortho}
\label{sec:method-intervention}
\label{sec:method-datasets}
\label{sec:method-gsrs}

\textbf{Impossibility direction.} At a fixed layer $L$ and under a class-stratified 50/50 held-out A/U (HO-AU) split, we fit PCA ($k{=}100$) on train-A states, project all states out of that A-subspace to a residual $R$, and estimate the train-split null-space mean difference
\[
\hat{d} = \frac{\mu_U^R - \mu_A^R}{\lVert \mu_U^R - \mu_A^R \rVert},
\]
and score test states by $\cos(R(x), \hat{d})$ using orientation-invariant AUC, averaged over 5 seeds per cell. This CosNSRT probe is a diagnostic instance of the Generalized Subspace Residual Score (GSRS), a Projection--Direction--Scoring template that also expresses Arditi's refusal direction. The earlier-grid GSRS ablation is shown in Fig.~\ref{fig:app-gsrs-ablation}, and the legacy layer-emergence analysis motivating layer selection is in Appendix~\ref{app:layers}.

\textbf{Safety refusal and orthogonality.} Following \citet{arditi2024refusal}, we construct $d_{\mathrm{ref,safety}} = \mu_{\mathrm{harmful}} - \mu_{\mathrm{harmless}}$ from 50 harmful and 50 harmless prompts at layer $L$, followed by behavior verification. We report $\cos(d_{\mathrm{imp}}, d_{\mathrm{ref,safety}})$ with $d_{\mathrm{imp}}$ in A-null and $d_{\mathrm{ref,safety}}$ in full space, together with bootstrap 95\% confidence intervals and a same-space A-null control.

\textbf{Steering and gated flip rate.} At layer $L$, a forward hook adds $\alpha \cdot \hat{d}$ to the residual stream during \texttt{model.generate}. We sweep $\alpha \in \{5, 10, 20, 40\} \cdot \sigma$ in both signs and compare against a random-direction control. The headline metric is \emph{gated flip rate}: the conditional probability of behavior change on samples whose clean baseline matches the pre-intervention class, under an invalidity-aware classifier with mixed-output and degenerate-output guards. The high-rigor v2 grid is 4 anchors (Mistral-7B-Instruct, Gemma-3-4B-it, Qwen3-14B, Qwen3-8B) $\times$ \{math800, code800, fact800\}; a 48-cell deterministic breadth sweep across 16 models supplies the across-grid check (Appendix~\ref{app:protocol-audit}). The datasets (math800 $16{\times}50$; code800 $8{\times}100$; fact800 800 SQuAD~2.0 pairs) are documented in Appendix~\ref{app:data}.

\textbf{Labeling protocol and provenance.} The v2 grid uses candidate labels assigned to all intervention records under a fixed written rubric through an LLM-assisted batch review, supported by deterministic domain-specific labeling utilities and schema/gate validation. A second LLM-assisted pass covered rubric-sensitive rows and stratified samples, and the first author reviewed uncertain cases; the first author did not independently review every row. For the nine non-Qwen3-8B cells, the aggregate JSONs apply candidate labels plus provisional second-pass audit fills; the three Qwen3-8B intervention cells use candidate-label passthrough with no second-pass override. We therefore describe the effective labels as LLM-assisted rather than human-adjudicated. The second pass identified over-credit in 10 of 37 provisional candidate flips on Mistral code, without changing that cell's best-dose effect of $+35.4$pp, and over-strict labeling in 2 of 300 checked rows on Gemma-3-4B fact. Among the 10 A$\to$U downshifts, mixed-output handling is the primary cause in 9 and degeneration contributes to 6, with overlap in 5. The degenerate-output guard changes Mistral fact A$\to$U at $\alpha{=}40$ from $+34$pp to $+4$pp. Gate broadening changes Mistral code U$\to$A from $+71$pp on a 14-row gate to $+44$pp on a 27-row gate. Counted strictly per slot, 17 of 24 effects decrease, 5 increase, 1 is unchanged, and 1 becomes unmeasurable; the earlier 18/4/2 summary uses a 6pp flatness convention (Appendix~\ref{app:protocol-audit}). Excluding the three Qwen3-8B candidate-only cells leaves the qualitative conclusion unchanged: Mistral-7B is the bidirectional structural keystone, while Gemma-3-4B and Qwen3-14B code remain positive in both directions.

\textbf{Artifacts and licenses.} Code, original \texttt{math800}/\texttt{code800} data, aggregate artifacts, and analysis scripts are released under MIT terms at \url{https://github.com/yucheng-du/recognition-refusal-misalignment}. \texttt{fact800} retains SQuAD~2.0's CC BY-SA 4.0 terms; the shipped AbstentionBench-GSM8K subset retains CC BY-NC 4.0 terms; the difficulty-controlled GSM8K derivative retains the upstream MIT terms. Because FalseQA has no explicit upstream license, our artifact does not redistribute it and instead provides a fetch-and-clean script subject to the upstream source terms.

Table~\ref{tab:claims-evidence} maps each claim to its evidence population and evidentiary role.

\begin{table*}[t]
\centering
\footnotesize
\setlength{\tabcolsep}{4pt}
\renewcommand{\arraystretch}{1.08}
\begin{tabular}{@{}c@{\hspace{5pt}}>{\raggedright\arraybackslash}p{0.28\textwidth}>{\raggedright\arraybackslash}p{0.46\textwidth}>{\raggedright\arraybackslash}p{0.19\textwidth}@{}}
\toprule
\# & Claim & Evidence & Tier \\
\midrule
1 & Recognition exists & 22-cell probe grid; HO-AU, 5-seed averages & headline \\ \addlinespace[2pt]
2 & Recognition is near-orthogonal to safety refusal & 22-cell geometry, same-space control, and energy decomposition & headline \\ \addlinespace[2pt]
3 & Recognition is partially aligned with behavior-defined invalidity awareness & Eight structural cells: six-cell primary summary plus two Qwen3-8B candidate-label cells & supporting \\ \addlinespace[2pt]
4 & Causal control of invalidity-aware behavior & Four-anchor v2 grid with LLM-assisted labels; Qwen3-8B candidate-only cells marked & headline for structural math/code; Mistral-7B is the bidirectional keystone \\ \addlinespace[2pt]
5 & Breadth across 16 models & 48-cell deterministic sweep & supporting only \\ \addlinespace[2pt]
6 & Geometry largely predates instruction tuning & Six math800 base/instruct pairs: five fully verified plus one proxy-base contrast & headline \\ \addlinespace[2pt]
7 & fact800 / FalseQA & Scoped boundary tests; fact800 also appears in the intervention grid, while FalseQA is zero-shot only & boundary; no headline claim \\
\bottomrule
\end{tabular}
\caption{Claims by evidence tier. Populations and metrics are not pooled across rows; in particular, the 48-cell breadth sweep uses a non-comparable deterministic metric and supports no per-cell claim.}
\label{tab:claims-evidence}
\end{table*}

\section{Core Findings}
\label{sec:findings}

\subsection{Recognition Exists}
\label{sec:findings-detection}

\textbf{Does a frozen LLM's internal state encode whether a structurally impossible question has no answer before any token is generated?}

\noindent Prior work documents at the output level that LLMs often proceed as if unreasonable math problems were well-posed \citep{ma2026umpr}. We test the internal claim directly in a narrower, formally verifiable structural-impossibility setting, with a deliberately low-capacity probe: one residual-stream direction, scored by cosine similarity, with no learned classifier on top.

A single A-null MeanDiff direction separates A from U prompts with mean AUC \textbf{0.939} across the 22-cell 11-model main grid, with range $[0.841, 0.993]$. Detection is not restricted to large models: SmolLM2-1.7B reaches 0.880 on math800; Qwen3-32B reaches 0.993; every 7B+ instruct model exceeds 0.90 on math800. Under identical MeanDiff plus cosine scoring, A-null projection is the load-bearing step (Table~\ref{tab:detection}). The geometric interpretation is that the top A-PCs capture what answerable prompts share, such as topic, surface form, and syntactic scaffolding, but not their answerability. Removing that variance exposes the impossibility-specific signal that is otherwise mixed with answerable-structure covariance.

\begin{table}[t]
\centering
\small
\setlength{\tabcolsep}{6pt}
\begin{tabular}{@{}lcc@{}}
\toprule
Subspace (MeanDiff $+$ cosine) & Mean AUC & Null $>$ this \\
\midrule
A-null ($d_{\mathrm{imp}}$) & \textbf{0.939} & --- \\
Full residual stream        & 0.907 & 19/22 \\
Top-$k$ A-PC                & 0.890 & 21/22 \\
\bottomrule
\end{tabular}
\caption{Detection of structural impossibility by a one-dimensional MeanDiff direction read under cosine similarity, across the 22-cell 11-model main grid (HO-AU, 5-seed averaged). A-null projection is the load-bearing factor: mean AUC \textbf{0.939} (range $[0.841, 0.993]$; SmolLM2-1.7B 0.880, Qwen3-32B 0.993), exceeding Full in 19/22 and Top-$k$ A-PC in 21/22 cells. Per-cell values in Fig.~\ref{fig:app-detection-heatmap} (AUC) and Fig.~\ref{fig:app-nullspace-ablation} (subspaces).}
\label{tab:detection}
\end{table}

We interpret the result as a linearly accessible pre-generation signal rather than merely an unconstrained probe prediction for two reasons. First, the probe is one-dimensional and read under cosine similarity: there is no parameter budget for a learned classifier to fit a complicated decision boundary, so what the probe recovers must be \textit{geometrically present} along a single direction in the residual stream. Second, the same classifier family fails when run in either the full residual stream or the top-$k$ A-PC subspace; only the A-null subspace exposes the signal cleanly. The information is in the model's representation, in a particular subspace, and a low-capacity reader is sufficient to recover it. A linear SVM in full-space is competitive on 3 of 4 representative cells (Appendix~\ref{app:svm}), so we frame A-null projection as an \textit{accessibility} result for low-capacity readers rather than as a claim that impossibility lives only in A-null; richer classifiers can navigate the full space too. This accessibility result licenses the rest of the paper: if the recognition signal is recoverable along a single direction, that direction is the natural object for geometric and causal questions.

\subsection{Recognition Is Not the Safety-Refusal Axis}
\label{sec:findings-ortho}

\begin{figure*}[t]
  \centering
  \includegraphics[width=0.85\textwidth]{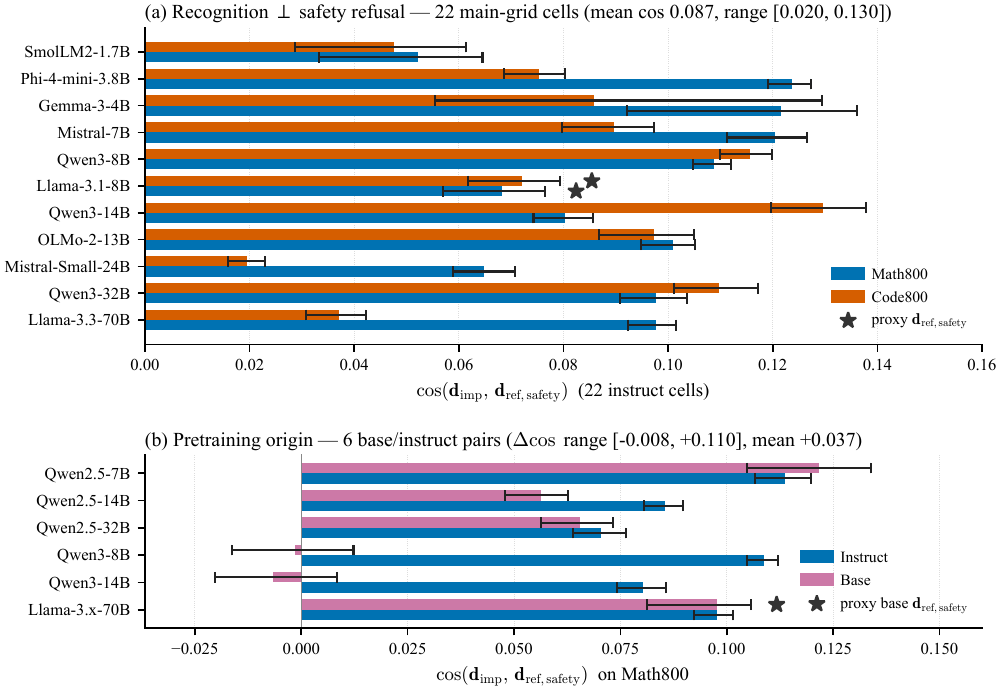}
  \caption{\textbf{Recognition is not the safety-refusal axis: near-orthogonality across the 11-model main grid and 6 base/instruct paired comparisons.} (a) 22 instruct cells (11 main-grid models $\times$ \{math800, code800\}) with bootstrap 95\% CIs; mean $\cos$ 0.087, range $[0.020, 0.130]$. Asterisks mark the two Llama-3.1-8B cells that use a behavior-verification proxy $d_{\mathrm{ref,safety}}$. (b) 6 math800 base/instruct paired comparisons (5 fully verified $+$ 1 Llama-3.1-70B proxy base, asterisked); $\Delta\cos$ range $[-0.008, +0.110]$, mean $+0.037$. The largest $\Delta\cos$ is the Qwen3-8B pair ($+0.110$); the Llama-3.3-70B-Instruct vs.\ Llama-3.1-70B-Base pair, the only vendor-confirmed post-training-only contrast, has $\Delta\cos \approx -0.0001$.}
  \label{fig:orthogonality}
\end{figure*}

If the model internally represents structural impossibility, why does it still answer? If recognition reused the trained safety-refusal pathway, $d_{\mathrm{imp}}$ should align with the canonical safety-refusal direction $d_{\mathrm{ref,safety}}$. We test this prediction directly.

For each model, we extract $d_{\mathrm{ref,safety}}$ from 50 harmful and 50 harmless prompts and apply behavior verification (\S\ref{sec:method}); 20 of 22 main-grid cells pass. The remaining two are the Llama-3.1-8B math800 and code800 cells, which use a proxy and are flagged in Fig.~\ref{fig:orthogonality}a. We compare $d_{\mathrm{ref,safety}}$ to $d_{\mathrm{imp}}$ at the matched layer and report $\cos(d_{\mathrm{imp}}, d_{\mathrm{ref,safety}})$ in the A-null subspace where the impossibility signal is read, with a 1{,}000-resample bootstrap confidence interval on each cell.

The headline number is mean cosine \textbf{0.087}, with range \textbf{$[0.020, 0.130]$}, across the 22 cells (Fig.~\ref{fig:orthogonality}a). The tightest bootstrap confidence interval is the 24B model on math800 (Mistral-Small-24B, $\cos = 0.065$, 95\% CI $[0.059, 0.071]$). Every interval lies in a near-orthogonal regime and excludes alignment. The cosines are small but not exactly zero: observed values are 2--13$\times$ the empirical random baseline. Thus, the two directions are neither aligned nor unrelated numerical noise; they have a consistent low-cosine relation.

A natural objection is that the cosine is measured across subspaces: $d_{\mathrm{imp}}$ is restricted to A-null, while $d_{\mathrm{ref,safety}}$ is read in the full residual stream. A same-space control rules this out. When both directions are projected into A-null before the cosine is taken, $\cos_{\mathrm{same},A\text{-null}}$ has range $[0.021, 0.154]$ and mean 0.097 across the 22 cells, comparable to the matched cosines. Equalizing the subspace does not remove the angle. The full-space cosine can be larger (22-cell range $[0.057, 0.781]$, mean 0.240), but energy decomposition shows why: across the 22 main-grid cells, the A-PC component accounts for mean 0.813 of the magnitude of $\cos_{\mathrm{full},\mathrm{full}}$ (range $[0.607, 0.983]$; Appendix~\ref{app:energy}). Both directions partly share A-PC variance, i.e., topic, format, and syntactic structure. Once we restrict attention to the A-null subspace where impossibility is accessible, recognition and safety refusal remain near-orthogonal. A four-cell $k\in\{5,10\}$ subspace analysis reaches the same low-overlap conclusion (Appendix~\ref{app:multidim}).

\begin{table}[t]
\centering
\footnotesize
\setlength{\tabcolsep}{3pt}
\begin{tabular}{@{}lccc@{}}
\toprule
Direction pair & $\cos$ & range & cells \\
\midrule
$d_{\mathrm{imp}}\cdot d_{\mathrm{ref,safety}}$        & 0.087 & $[0.020, 0.130]$ & 22 main \\
$d_{\mathrm{imp}}\cdot d_{\mathrm{struct,behav}}$      & \textbf{0.40} & $[0.16, 0.59]$ & 6 primary \\
$d_{\mathrm{struct,behav}}\cdot d_{\mathrm{ref,safety}}$ & 0.08 & --- & 6 primary \\
\bottomrule
\end{tabular}
\caption{Pairwise geometry of the three abstention-related directions (matched layer). Recognition is near-orthogonal to the canonical safety-refusal axis (0.087 across the 22-cell main grid), only partially aligned with the in-domain invalidity-aware behavior direction (0.40), which is itself near-orthogonal to safety refusal (0.08). \emph{Populations differ}: rows 2--3 summarize the six-cell primary non-Qwen3-8B behavior subset (8-cell mean including 2 Qwen3-8B candidate-only cells: 0.38 / 0.077), on which $\cos(d_{\mathrm{imp}}, d_{\mathrm{ref,safety}}) = 0.13$, not the 22-cell 0.087. Controls (22-cell): projecting both directions into A-null leaves the angle near-orthogonal ($\cos$ 0.097); the larger full-space $\cos$ (0.240) is 0.813 shared A-PC (answerable-structure) variance. Per-cell values in App.~\ref{app:struct-behav}, \ref{app:energy}.}
\label{tab:geometry}
\end{table}

\textbf{Behavior-defined direct comparison.} Because $d_{\mathrm{ref,safety}}$ is built from harmful-vs-harmless prompts \citep{arditi2024refusal}, a low $\cos(d_{\mathrm{imp}}, d_{\mathrm{ref,safety}})$ could merely mean that harmfulness and structural impossibility have different prompt form. To separate this surface-form issue from behavior, we construct an in-domain behavior-defined invalidity-aware direction $d_{\mathrm{struct,behav}}$. This direction contrasts U-class clean-baseline generations that the model labels invalidity-aware against U-class generations that answer anyway (construction, bootstrap confidence intervals, the one exploratory cell, and full-space robustness are in Appendix~\ref{app:struct-behav}). Across the four-anchor $\times$ \{math800, code800\} grid, recognition aligns more with the behavior-defined direction than with safety refusal, but only partially; the behavior-defined direction is itself near-orthogonal to safety refusal (Table~\ref{tab:geometry}). Strict refusal-only generations are essentially absent (0 of 50 U prompts per cell), so a natural strict-refusal direction is not constructible. This absence is itself evidence that models rarely produce standard refusal language for structurally impossible prompts, even when they recognize invalidity. These comparisons do not exhaust the model's abstention mechanisms; unmeasured directions could mediate other abstention routes.

$d_{\mathrm{ref,safety}}$ carries some A/U predictive power on math800 (11-cell range $[0.600, 0.962]$, mean 0.834), as expected from shared answerable-structure overlap: any direction with nontrivial A-PC energy can pick up surface variance that separates A and U prompts. But the A-null component of $d_{\mathrm{imp}}$, the component that carries the structural-impossibility signal, is not the axis read by the safety-refusal mechanism.

The confident-on-impossible failure is therefore not caused by a missing abstention-related representation. Recognition exists and is partially aligned with in-domain invalidity-aware behavior. However, that behavior direction is near-orthogonal to the trained safety-refusal channel, and explicit refusal language is nearly absent. The recognition signal is present; the trained refusal route is not aligned with it.

\subsection{The Recognition Direction Causally Controls Invalidity-Aware Behavior}
\label{sec:findings-causal}

\begin{figure}[t]
  \centering
  \includegraphics[width=\columnwidth]{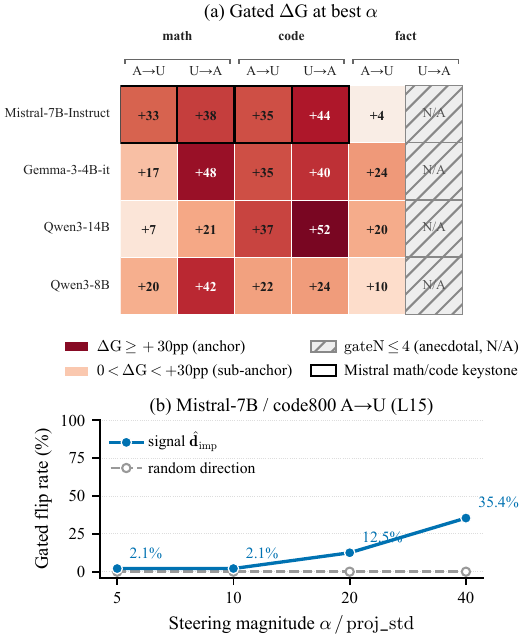}
  \caption{\textbf{The recognition direction causally controls invalidity-aware behavior.} (a)~Gated $\Delta$G at the best steering magnitude $\alpha$ across the 4-anchor intervention grid (Mistral-7B-Instruct, Gemma-3-4B-it, Qwen3-14B, Qwen3-8B) $\times$ \{math, code, fact\}, both directions: darker cells are anchor-quality ($\Delta$G $\geq +30$pp), the boxed Mistral-7B math/code cells are the bidirectional keystone, and hatched cells are anecdotal (gateN $\leq 4$, all U$\to$A on fact). (b)~The Mistral-7B code A$\to$U dose-response: gated flip rate rises monotonically with $\alpha$ ($2.1\to2.1\to12.5\to35.4$\%) against a near-zero random-direction baseline. See \S\ref{sec:findings-causal}.}
  \label{fig:causal}
\end{figure}

A direction can correlate with answerability without driving generation behavior; probing alone cannot distinguish these cases \citep{belinkov2022probing,ravichander2021probing}. We therefore ask whether the recognition direction is a causal handle. If $d_{\mathrm{imp}}$ is behaviorally active, then adding it during generation should increase invalidity-aware behavior on A prompts, subtracting it should reduce invalidity-aware behavior on U prompts that already abstain, and a random direction at matched magnitude should not show the same effect.

\textbf{Protocol.} We test on a 4-anchor intervention grid: Mistral-7B-Instruct, Gemma-3-4B-it, Qwen3-14B, Qwen3-8B $\times$ \{math800, code800, fact800\}. At each model--dataset cell, we register a forward hook at the matched layer that adds $\alpha \cdot \hat{d}$ to the residual stream at the last-token position of every forward pass during \texttt{model.generate}. We sweep $\alpha \in \{5, 10, 20, 40\} \cdot \sigma$ in both signs, separately for $\hat{d} = d_{\mathrm{imp}}$ and for a random unit direction. In the U$\to$A condition, we subtract $d_{\mathrm{imp}}$ from U prompts; in the A$\to$U condition, we add $d_{\mathrm{imp}}$ to A prompts. The headline metric is gated flip rate, defined in \S\ref{sec:method}.

\textit{Mistral-7B is the keystone causal anchor.} Mistral-7B is the only model in the 4-anchor grid whose intervention exceeds $+30$pp gated $\Delta$G in both directions and in both structural domains: math A$\to$U $+33$pp, math U$\to$A $+38$pp, code A$\to$U $+35$pp, and code U$\to$A $+44$pp (Fig.~\ref{fig:causal}a). At the best operating points, the signal direction exceeds its matched random-direction control by $+33$ to $+44$pp gated. The cleanest dose response is Mistral code A$\to$U, where the gated flip rate increases monotonically across $\alpha=5\to10\to20\to40$ with a near-zero random baseline (Fig.~\ref{fig:causal}b).

\textit{Other anchors are direction-asymmetric or domain-specific.} Code remains anchor-quality in both directions for Gemma-3-4B ($+35$/$+40$pp) and Qwen3-14B ($+37$/$+52$pp); Qwen3-8B is positive but sub-anchor on code ($+22$/$+24$pp). Math control is direction-asymmetric: Gemma-3-4B and Qwen3-8B reach anchor-quality on U$\to$A only ($+48$pp and $+42$pp), while their A$\to$U directions degenerate at higher $\alpha$; Qwen3-14B math fails in both directions. Of 24 anchor--dataset--direction slots, 10 are anchor-quality, 10 are positive but sub-anchor, and 4 are anecdotal (gateN $\leq 4$, all U$\to$A fact).

\textit{Why effects vary across models.} Post hoc diagnostics are consistent with a two-factor account in which steering succeeds when the dose required to flip behavior lies inside the model's tolerance window for residual-stream perturbation. All 10 anchor-quality structural directions reach $+30$pp with signal-branch degeneration no higher than $16\%$; five of the six sub-threshold directions reach at least $40\%$ degeneration at a tested dose, while Qwen3-8B code A$\to$U remains below $21\%$ and peaks at $+22$pp. At tested doses with under $25\%$ degeneration, all six sub-threshold directions have a positive best signal-minus-random effect ($+7$ to $+24$pp). At $\alpha{=}20$, mean structural-cell degeneration is $9.5\%$ for Mistral-7B versus $31.5$--$45.5\%$ for the other anchors; matched-norm random directions reproduce the Mistral-versus-rest tolerance gap. Recognition-to-behavior coupling is descriptively associated with the minimum U$\to$A anchor dose (Spearman $\rho=-0.86$ across eight structural cells), but not with best effect size; across the four cells that reach the anchor in both directions, A$\to$U requires 2--4 times the U$\to$A dose. This account is post hoc and correlational: the eight cells come from four anchors in three model families, and tolerance, matched-layer depth, and Qwen3 post-training are confounded (Appendix~\ref{app:protocol-audit-dose}).

\textit{fact800 is an epistemic transfer boundary.} On fact800, the U$\to$A direction is not reliably measurable: clean U baselines almost never abstain in passage-grounded language (gateN 0, 1, 4, 2 across the four anchors), so flip rates are small-$N$ anecdotes (Appendix~\ref{app:protocol-audit-anecdotal}). A$\to$U is well measured (gateN $\approx 49$--$50$) but uniformly sub-anchor ($+4$ to $+24$pp). This boundary reflects the absence of a passage-grounded abstention baseline on SQuAD-style unanswerability, not necessarily the absence of any fact-domain recognition effect.

\textit{Refusal-only universality.} Of 96 non-baseline rows (cell $\times$ direction $\times$ $\alpha$), 48 are unmeasurable under a refusal-only criterion because clean baselines almost never produce explicit ``I cannot'' or ``I do not know'' phrasing. Of the 48 measurable rows, 47 have refusal-only gated $\Delta$G $\leq +5$pp. The lone exception is Mistral math A$\to$U at $\alpha{=}20$, with $+6.0$pp, an order of magnitude below the matched invalidity-aware $\Delta$G of $+32.7$pp. These results show that $d_{\mathrm{imp}}$ is not simply a generic refusal-vocabulary axis.

\textit{Steering breadth across 48 cells.} A coarser deterministic breadth sweep across 16 models $\times$ 3 datasets reproduces the math/code footprint and confirms fact800 as the weakest domain (per-domain hallucination-reduction numbers in Appendix~\ref{app:protocol-audit}). Because this sweep scores hallucination-rate reduction rather than signal-minus-random gated flips, it is not a comparable effect size to intervention $\Delta$G, ranks math versus code differently, and supports no per-cell claim. All headline causal numbers come from the 4-anchor intervention grid.

Taken together with \S\ref{sec:findings-ortho}, the intervention result strengthens the routing account. Geometry shows that recognition is near-orthogonal to safety refusal. Steering shows that injecting along the recognition axis changes invalidity-aware behavior in the structural setting where the geometric measurement is cleanest. The two together rule out an epiphenomenal reading: $d_{\mathrm{imp}}$ is a behaviorally active handle, but the safety-refusal mechanism does not read that axis. That is the mechanistic restatement, in our formal structural setting, of the output-level unreasonable-math failure documented by \citet{ma2026umpr}.

\subsection{The Misalignment Largely Predates Alignment}
\label{sec:findings-pretrain}

The angle in \S\ref{sec:findings-ortho} could be created by post-training. Instruction tuning and preference optimization install the safety-refusal mechanism, and these stages might place it on an axis unrelated to structural-impossibility recognition because safety rewards do not contain structural-impossibility labels. The competing account is that the two directions are already near-perpendicular at the pretraining endpoint, and post-training turns a harmful-vs-harmless statistical direction into an active refusal mechanism without changing its relation to $d_{\mathrm{imp}}$ very much. These accounts make different predictions for $\cos(d_{\mathrm{imp}}, d_{\mathrm{ref,safety}})$ on base models, so paired base/instruct measurements are decisive. This comparison concerns the geometry between recognition and the canonical safety-refusal direction; it does not cover the in-domain invalidity-aware behavior direction from \S\ref{sec:findings-ortho}, whose construction requires instruction-following behavior labels that are unavailable for base models.

We compare $d_{\mathrm{imp}}$ and $d_{\mathrm{ref,safety}}$ across 6 paired base/instruct checkpoints on math800 at matched layers. The pair list, per-pair verification status, and sourcing caveats, including the Llama-3.1-70B proxy base and the Qwen2.5-32B scale substitution for the unreleased Qwen3-32B base, are in Appendix~\ref{app:struct-behav}. The matched-layer $\Delta\cos$ (instruct minus base) ranges $[-0.008, +0.110]$ with mean $+0.037$ (Fig.~\ref{fig:orthogonality}b), small relative to the 22-cell instruct band $[0.020, 0.130]$. Within-family pairs track closely. The largest shift is Qwen3-8B ($+0.110$), and the vendor-confirmed post-training-only contrast, Llama-3.3-70B-Instruct versus Llama-3.1-70B-Base, has $\Delta\cos \approx -0.0001$. The strongest evidence is the five fully verified Qwen pairs; the Llama-70B pair, which uses a proxy base, is a complementary controlled contrast.

The 6 pairs support the same conclusion. Instruction tuning changes the angle in a family-dependent way within a low-cosine regime, but it does not create the regime, and none of the 6 pairs closes the angle. The near-perpendicular geometry is already present at the pretraining endpoint. Recognition itself is also present before instruction tuning: a one-dimensional A-null probe detects impossibility on base-model math800 with mean AUC $\approx 0.98$ across the six paired base checkpoints (range $[0.94, 0.99]$; per-model values in Appendix~\ref{app:struct-behav}). Thus, both the recognition signal and its separation from safety refusal largely predate instruction tuning.

\section{Ruling Out Alternatives and Mapping Scope}
\label{sec:char}
\label{sec:char-scaling}
\label{sec:char-transfer}
\label{sec:char-length}
\label{sec:char-difficulty}
\label{sec:char-form}

\begin{table}[t]
\centering
\footnotesize
\setlength{\tabcolsep}{5pt}
\begin{tabular}{@{}lp{2.05in}@{}}
\toprule
Ruled out & Key evidence (per-cell detail in App.~\ref{app:robustness}) \\
\midrule
Length & Length-matched GSM8K (length-only AUC ${\to}$ 0.500): CosNSRT 0.799 \\
Difficulty & Hard-vs-easy 0.61 vs.\ impossibility 0.96 ($\Delta {=} {+}0.35$) \\
Category & NS\_SNR--AUC Spearman $\rho$ 0.730 (18/22 cells $p {<} 0.05$) \\
Scale & AUC 0.841--0.993; 24B $\cos$ 0.065 [0.059,\,0.071]; 32B/70B ${\approx}$ 0.098 \\
Generic axis & Within$-$cross drop 0.080; dot-product NSRT transfer 0.64--0.98 (GSM8K), 0.59--0.90 (FalseQA) \\
\bottomrule
\end{tabular}
\caption{Five alternative explanations ruled out for the \S\ref{sec:findings} findings, each with its key result. Structural impossibility (math800, code800) is the clean setting; fact800 and FalseQA are boundary cases, not equivalent evidence. Full per-control evidence in App.~\ref{app:robustness}.}
\label{tab:robustness}
\end{table}

\looseness=-1 The four findings in \S\ref{sec:findings} survive five alternative explanations and have a clear scope boundary. Table~\ref{tab:robustness} pairs each control with its key result; full per-cell evidence is in Appendix~\ref{app:robustness}. The recognition signal is not explained by prompt length, answerable-class difficulty, a few easy categories, scale, or a generic unanswerability axis. Structural math and code remain the main setting; fact800 and FalseQA are transfer boundaries.

\section{Related Work}
\label{sec:related}
\textbf{Unanswerability regimes and abstention.} Questions for which answer-like completion is inappropriate span distinct regimes that we deliberately keep apart (\S\ref{sec:background}). \emph{Epistemic unanswerability} covers cases where a correct answer may exist but is not recoverable from the provided evidence, as exemplified by SQuAD~2.0 \citep{rajpurkar2018squad2}. \emph{False-premise} questions instead presuppose a false fact and should be challenged rather than answered \citep{kim2021presupposition,hu2023falseqa}. \emph{Structural impossibility}, our setting, is narrower still: a formal rule makes \emph{no} admissible answer exist; \citet{ma2026umpr} document that LLMs nonetheless treat unreasonable math problems as well-posed, and our code setting extends this question beyond prior math-focused evidence. Cutting across these regimes, work on selective question answering, dedicated unanswerability benchmarks, and abstention surveys asks when models should decline, abstain, or say ``I don't know'' \citep{kamath2020selective,feng2025abstentionbench,wen2025know}. Much of that literature is diagnostic and measured at the output level; we instead ask, in the structural regime where ground truth is rule-verifiable, whether the ``no answer'' signal is present internally before generation, whether it coincides with the canonical safety-refusal axis, and whether it causally drives invalidity-aware abstention.

\looseness=-1 \textbf{Pre-generation probes and abstention.} Representation-probing work reads model internals for truthfulness, hallucination, or latent-knowledge signals \citep{orgad2025llms,burns2023discovering}, while FacLens explicitly predicts future non-factual QA responses from hidden question representations before generation \citep{chen2025faclens}; \citet{levinstein2023still} caution that such truthfulness probes face empirical and conceptual generalization limits. We share the internal-readout stance but target a narrower, formally verifiable property (structural impossibility), read it from an A-null subspace, and add causal validation by asking whether the probed feature \emph{drives} behavior, a question that line leaves open. \citet{kim2025detecting} detect hallucination from layer-wise information deficiency on ambiguous and unanswerable prompts, and \citet{du2026geometric} uses geometric deviation as an unsupervised answerability signal, whereas we ask whether a null-space recognition direction aligns with the canonical safety-refusal axis or causally drives invalidity-aware abstention.

\textbf{Refusal geometry and steering.} Our closest neighbor is \citet{arditi2024refusal}, whose MeanDiff $d_{\mathrm{ref,safety}}$ we adopt and behavior-verify; \citet{wollschlager2025geometry} refine refusal geometry into concept cones, \citet{latentbiopsy2026} casts harmful-intent detection as angular deviation from safe-prompt residual geometry, and \citet{wu2026knowing} decompose safety refusal into recognition and execution axes \emph{within} that pathway. We instead measure the angle \emph{between two distinct abilities} (impossibility recognition and safety refusal) and find it already present at pretraining (\S\ref{sec:findings-pretrain}). Our steering protocol follows \citet{turner2023activation,li2023inference} (closest to \citealp{arditi2024refusal}); unlike strands that steer to harden safety or boost tasks \citep{liu2026adaras,lee2025cast,zhang2025safeswitch}, we use it only to test whether $d_{\mathrm{imp}}$ is a causal handle on abstention.

\section{Discussion \& Conclusion}
\label{sec:discussion}
The confident-on-impossible failure in our structural math/code settings is better explained as a routing failure than as an encoding failure. The model contains the information needed to abstain (\S\ref{sec:findings-detection}), but that information is near-orthogonal to the canonical safety-refusal axis, only partially coupled to an in-domain invalidity-aware behavior axis, and accompanied by essentially no strict refusal-only language (\S\ref{sec:findings-ortho}). This geometry predates instruction tuning (\S\ref{sec:findings-pretrain}); steering changes invalidity-aware behavior on structural math/code cells, while fact800 remains a boundary and refusal-only scoring rules out a generic refusal-vocabulary reading (\S\ref{sec:findings-causal}). The trained safety-refusal pathway reads a different axis. Read through our results, the descriptive ``unconscious of unreasonableness'' framing \citep{ma2026umpr} is about routing, not encoding. The auxiliary AbstentionBench-GSM8K and FalseQA results are detection-only boundary tests under dot-product NSRT or CosNSRT scoring, not evidence that steering transfers beyond the controlled structural setting.

The same recognition-action separation may recur when epistemic confidence, harmfulness, or instruction-compliance signals fail to drive hedging, refusal, or format adherence. A-null probing, behavior-verified MeanDiff refusal directions, matched-layer cosine, generation-time gated steering, and base/instruct contrasts can test such cases directly. To close the angle, future work could train the safety-refusal pathway to read $d_{\mathrm{imp}}$ without collapsing it into $d_{\mathrm{ref,safety}}$; the Llama-70B contrast shows that controlled post-training can leave the angle essentially unchanged ($\Delta\cos \approx -0.0001$), so direct structural-impossibility supervision may be needed. Models represent impossibility before generation, but that recognition is not reliably routed into abstention.

\section*{Limitations}
\label{sec:limitations}
\looseness=-1 Our causal evidence is cleanest on \textit{structural} impossibility. Code shows anchor-quality control on 3/4 anchors; math control is direction-asymmetric on the non-Mistral anchors with Qwen3-14B $\times$ math the weakest causal cell under v2; fact800 is a structural boundary case rather than a clean replication. The clean causal regularity is therefore a property of structural impossibility rather than of unanswerability in general; readers should interpret each finding as scoped to the setting it was measured in. Safety refusal and the measured behavior-defined direction are two comparators, not an exhaustive basis for abstention: unmeasured directions could mediate other abstention routes. The main analysis uses single directions; a four-cell 5--10-dimensional linear-subspace check preserves the low-overlap conclusion (Appendix~\ref{app:multidim}), while nonlinear or more distributed representations remain open. CosNSRT fixes PCA $k=100$ as a heuristic and does not tune it per model or dataset; the subspace ablation isolates the projection as the load-bearing factor, but the exact $k$ is a design choice. The pipeline requires labeled A-class prompts to fit the null-space basis, which is available for math800 / code800 but may be costly to construct on new domains.

\textbf{Architecture scope.} The 11-model main grid evaluates open-weight transformer-family language checkpoints in a text-only setting and analyzes the residual stream of the language decoder. This is a deliberate scope choice rather than a claim about all contemporary model families. Some included checkpoints are distributed in multimodal-capable families; we use no image inputs and make claims only about their text decoder path. Our geometric analysis assumes that abstention-relevant features can be read as single residual-stream directions, an assumption well supported for refusal in decoder-only language models \citep{arditi2024refusal} but not yet established for settings with vision-token conditioning, mixture-of-experts routing, hybrid attention, or explicit reasoning-mode training. Whether recognition $\perp$ safety refusal persists in those settings, and whether generation-time steering still yields clean dose-responsive control, requires separate validation and is left to future work.

Headline flip-rate metrics rely on a per-output classifier; v2 verification under an invalidity-aware rubric (Appendix~\ref{app:protocol-audit}) supersedes the v1 keyword detector, and the protocol-refinement audit documents per-cell shifts and the rationale for each. The high-rigor v2 intervention grid is 4 anchors (Mistral-7B-Instruct, Gemma-3-4B-it, Qwen3-14B, Qwen3-8B), not the full 11-model main grid; broader behavioral evidence comes from the 48-cell v2-deterministic steering breadth sweep across 16 models $\times$ 3 datasets, which uses a coarser regex-based proxy aligned to the intervention rubric (Appendix~\ref{app:protocol-audit}). Qwen3-32B has no public base release as of May 2026, so the 32B base/instruct paired comparison uses Qwen2.5-32B at the 32B scale rather than Qwen3-32B; the within-grid scope split is deliberate. The Llama-70B base side is a proxy $d_{\mathrm{ref,safety}}$ because Llama-3.1-70B-Base does not refuse harmful prompts often enough to behavior-verify; Meta confirms Llama-3.3-70B-Instruct shares the Llama-3.1-70B pretraining checkpoint, which makes the Llama-3.3 vs.\ Llama-3.1-70B pair the only vendor-confirmed post-training-only contrast in the grid. We report this pair as such, but explicitly do \textit{not} call it the ``cleanest evidence overall'': the proxy base $d_{\mathrm{ref,safety}}$ is a real caveat, and the strongest within-family evidence in \S\ref{sec:findings-pretrain} comes from the five fully-verified Qwen pairs.

\section*{Ethical Considerations}
Improved recognition-to-abstention routing could reduce confidently wrong outputs on structurally impossible requests. The same diagnostics and steering methods could also be misused to suppress warranted abstention or otherwise alter refusal behavior without users' knowledge. Any deployment should therefore evaluate both erroneous answering and erroneous abstention, preserve user-visible uncertainty, and restrict intervention controls to authorized settings.

\section*{Acknowledgments}
We thank the reviewers and area chair for their constructive feedback. We used AI assistants for writing, coding, and verification, as well as for the LLM-assisted dataset generation and labeling described in the appendices. The authors reviewed and approved all scientific decisions and final paper content and take responsibility for the paper.

\bibliography{references}

\appendix

\section{Data Construction}
\label{app:data}
This appendix documents the construction of the three study datasets (\texttt{math800}, \texttt{code800}, \texttt{fact800}) and the cleaning of the external \texttt{falseqa} file used for zero-shot transfer. The \texttt{math800} and \texttt{code800} prompts were generated with an API-hosted LLM under fixed category-specific instructions and then subjected to the deterministic validity and deduplication checks described below; the released scripts record the generation configuration. \texttt{fact800} and \texttt{falseqa} do not use LLM generation. Per-category probe AUC for \texttt{math800} and \texttt{code800} is reported separately in App.~\ref{app:per-category-auc}.

\subsection{math800}
\label{app:data-math}

\textbf{Scale.} 16 categories $\times$ 50 matched A/U pairs $=$ 1600 prompts (800 A, 800 U).

\textbf{Goal and U definition.} Each pair targets the \emph{same underlying mathematical operation or concept} and differs only in answerability under standard mathematics. An A prompt is answerable under standard mathematics; a U prompt is structurally unanswerable, restricted to four types: (i) formally undefined operations (e.g., $\div 0$, modulo by zero, invalid log base), (ii) non-existent results (e.g., derivatives at non-differentiable points, limits that do not exist), (iii) no solution in the stated domain (e.g., $x^2 = -4$ over $\mathbb{R}$), and (iv) divergent quantities. Real-world unknown quantities, future values, very large but computable numbers, $0^0$, and ``largest prime'' are explicitly excluded because they are answerable under standard conventions.

\textbf{Prompt format.} Every prompt is prefixed with \texttt{"Answer concisely: "} followed by a math question with varied surface templates.

\begin{table*}[t]
\centering
\small
\setlength{\tabcolsep}{5pt}
\begin{tabular}{@{}llll@{}}
\toprule
\textbf{Category} & \textbf{Code} & \textbf{Unanswerability type} & \textbf{Example U prompt} \\
\midrule
\texttt{div\_zero} & \texttt{mdiv} & Division by zero & $84 \div 0$ \\
\texttt{mod\_zero} & \texttt{mmod} & Modulo by zero & $23 \bmod 0$ \\
\texttt{sqrt\_negative\_real} & \texttt{msqrt} & $\sqrt{x}$ for $x<0$ in $\mathbb{R}$ & $\sqrt{-9}$ in the reals \\
\texttt{even\_root\_negative} & \texttt{meven} & Even root of negative in $\mathbb{R}$ & $\sqrt[4]{-16}$ in the reals \\
\texttt{neg\_base\_frac\_exp} & \texttt{mneg} & Negative base with fractional exponent in $\mathbb{R}$ & $(-8)^{1/2}$ in $\mathbb{R}$ \\
\texttt{log\_base\_zero\_or\_neg} & \texttt{mlog} & Log with invalid base & $\log_0(5)$, $\log_{-2}(8)$ \\
\texttt{factorial\_invalid} & \texttt{mfact} & Non-integer or negative factorial & $(-3)!$, $(2.5)!$ \\
\texttt{trig\_undefined} & \texttt{mtrig} & Trig function at undefined point & $\tan(\pi/2)$, $\cot(0)$ \\
\texttt{inv\_trig\_out\_of\_domain} & \texttt{minv} & Inverse trig out of domain in $\mathbb{R}$ & $\arcsin(2)$, $\arccos(-3)$ \\
\texttt{no\_real\_solution} & \texttt{mno\_r} & Equation with no real solution & $x^2 = -5$ in $\mathbb{R}$ \\
\texttt{gcd\_lcm\_irrational} & \texttt{mgcd} & GCD/LCM involving irrational & $\gcd(6, \pi)$ \\
\texttt{invalid\_combination} & \texttt{minva} & $C(n,k)$ with $k>n$ or $k<0$ & $C(3, 10)$, $C(6, -3)$ \\
\texttt{divergent\_series} & \texttt{mdive} & Sum of divergent series & $\sum_{n=1}^{\infty} n^2$ \\
\texttt{nonexistent\_limit} & \texttt{mnone} & Limit does not exist & $\lim_{x\to 0} \sin(1/x)$ \\
\texttt{singular\_matrix\_inverse} & \texttt{msing} & Inverse of singular matrix & $\bigl[\begin{smallmatrix}2 & 4 \\ 1 & 2\end{smallmatrix}\bigr]^{-1}$ \\
\texttt{undefined\_derivative} & \texttt{muder} & Derivative at non-differentiable point & $\tfrac{d}{dx}|x|$ at $x=0$ \\
\bottomrule
\end{tabular}
\caption{The 16 \texttt{math800} categories with representative unanswerable prompts. The ``Code'' column is the short identifier used in Appendix~\ref{app:per-category-auc} (Table~\ref{tab:per-cat-math}) for per-category AUC.}
\label{tab:math800-categories}
\end{table*}

\textbf{Anti-confound measures.}
\begin{enumerate}[leftmargin=*,itemsep=2pt,topsep=2pt]
\item \emph{Matched by operation / concept.} Each A/U pair addresses the same underlying mathematical operation with a matched surface template.
\item \emph{Structural U definition.} U prompts are restricted to the four structural types above; ``unanswerable because the model doesn't know'' and ``unanswerable because the number is big'' are not accepted.
\item \emph{Template diversity with short-answer control.} All prompts carry the \texttt{"Answer concisely: "} prefix while surface forms vary within category (\texttt{What is} / \texttt{Compute} / \texttt{Calculate} / \texttt{Find} / \texttt{Evaluate} / \texttt{Determine}, etc.) to reduce brittle template shortcuts.
\item \emph{Length confound control.} Token-level length AUC (Mistral tokenizer, $U = 1$) is $0.498$ globally, and all 16 categories fall within $[0.47, 0.53]$; no category approaches a severe length-confound regime.
\item \emph{Generator dedup guardrail.} \texttt{src/data/generate\_math800.py} rejects whitespace-insensitive duplicate prompts and, in \texttt{--resume} mode, seeds its dedup set from on-disk prompts in the target file rather than only from the current batch.
\end{enumerate}

\textbf{Audit status (2026-04-12).} 1600 rows, 16 categories $\times$ 50 matched pairs, zero duplicate ids, zero raw duplicate prompts, zero whitespace-insensitive duplicate prompts, zero incomplete pairs. Earlier versions contained 16 mislabeled \texttt{nonexistent\_limit} A prompts and 3 malformed or differentiable \texttt{undefined\_derivative} U prompts; these are fixed in the shipped file, and the shipped \texttt{undefined\_derivative} family is restricted to four accepted U types (absolute-value cusps, step discontinuities at half-integer ties, $\mathrm{sign}(x)$ at $x=0$, and $x^{1/3}$ at $x=0$). Exact file reproduction is not claimed because the API-based generator is stochastic.

\subsection{code800}
\label{app:data-code}

\textbf{Scale.} 8 categories $\times$ 100 matched A/U pairs $=$ 1600 prompts.

\textbf{Goal and U definition.} Each pair targets the \emph{same Python operation or API family}. An A prompt is an expression that returns a value in standard CPython; a U prompt raises a category-correct runtime exception or never terminates.

\begin{table*}[t]
\centering
\small
\setlength{\tabcolsep}{6pt}
\begin{tabular}{@{}llll@{}}
\toprule
\textbf{Category} & \textbf{Code} & \textbf{Error type} & \textbf{Example U} \\
\midrule
\texttt{zero\_division} & \texttt{czero} & \texttt{ZeroDivisionError} & \texttt{1 / 0} \\
\texttt{type\_error} & \texttt{ctype} & \texttt{TypeError} & \texttt{'hello' + 5} \\
\texttt{value\_error} & \texttt{cvalu} & \texttt{ValueError} & \texttt{int('abc')} \\
\texttt{index\_key\_error} & \texttt{cinde} & \texttt{Index}/\texttt{KeyError} & \texttt{[1,2,3][10]} \\
\texttt{attribute\_error} & \texttt{cattr} & \texttt{AttributeError} & \texttt{(5).append(1)} \\
\texttt{name\_error} & \texttt{cname} & \texttt{NameError} & \texttt{undef + 1} \\
\texttt{math\_domain\_error} & \texttt{cmath} & \texttt{math} domain & \texttt{math.sqrt(-1)} \\
\texttt{infinite\_iter} & \texttt{cinfi} & Non-termination & \texttt{list(count())} \\
\bottomrule
\end{tabular}
\caption{The 8 \texttt{code800} categories. The ``Code'' column is the short identifier used in Appendix~\ref{app:per-category-auc} (Table~\ref{tab:per-cat-code}) for per-category AUC.}
\label{tab:code800-categories}
\end{table*}

\textbf{Anti-confound measures.}
\begin{enumerate}[leftmargin=*,itemsep=2pt,topsep=2pt]
\item \emph{Matched by operation / function.} Each A/U pair targets the same operation family, differing only in whether the expression is well-formed for CPython.
\item \emph{Category-pure U semantics.} U expressions raise category-correct runtime outcomes only; \texttt{infinite\_iter} uses genuinely non-terminating iterators built from \texttt{itertools.count}, \texttt{itertools.cycle}, and \texttt{itertools.repeat}, consumed by operations such as \texttt{sum}, \texttt{list}, \texttt{tuple}, \texttt{max}, \texttt{min}, \texttt{sorted}, \texttt{set}.
\item \emph{Template diversity with short-answer control.} All prompts use the \texttt{"Answer concisely: "} prefix and one of 7 canonical templates with backtick-enclosed expressions (e.g.\ \texttt{``What is the result of evaluating `X` in Python?''}).
\item \emph{Length confound control.} Global token-level length AUC is $0.498$; all 8 categories fall in $[0.47, 0.53]$.
\item \emph{AST-level dedup.} \texttt{src/data/generate\_code800.py} rejects AST-equivalent duplicate expressions (not merely exact-string duplicates) and in \texttt{--resume} mode seeds its AST-dedup set from on-disk prompts.
\end{enumerate}

\textbf{Audit status (2026-04-12).} 1600 rows, 8 categories $\times$ 100 matched pairs, zero duplicate ids, zero raw duplicate prompts, zero AST-normalized duplicate expression groups, zero incomplete pairs. All non-\texttt{infinite\_iter} A expressions evaluate successfully under \texttt{math, itertools}; all corresponding U expressions raise category-correct exceptions; all \texttt{infinite\_iter} U prompts remain in genuinely non-terminating families. Exact file reproduction is not claimed because the API-based generator is stochastic.

\subsection{fact800}
\label{app:data-fact}

\textbf{Scale and source.} 800 matched pairs $=$ 1600 prompts, drawn from the SQuAD~2.0 \emph{train} split.

\textbf{Goal.} A natural epistemic-unanswerability control set in which answerability differs only at the QA level, not at the passage level. Each pair uses the same paragraph context and consists of an answerable SQuAD question (\texttt{is\_impossible=False}) and an unanswerable one (\texttt{is\_impossible=True}).

\textbf{Prompt format.} \texttt{Context: $\langle$truncated paragraph$\rangle$ / Question: $\langle$question$\rangle$ / Answer:} (newline-separated).

\textbf{Anti-confound measures.}
\begin{enumerate}[leftmargin=*,itemsep=2pt,topsep=2pt]
\item \emph{Matched by paragraph.} A and U in each pair come from the same SQuAD paragraph, so passage topic, style, and world knowledge are controlled within pair.
\item \emph{Verbatim-prefix context truncation.} Paragraph context is truncated at a 150-word boundary using the \emph{verbatim original prefix}, not a whitespace-normalized reconstruction; when truncation applies, the suffix \texttt{" ..."} is appended.
\item \emph{Article diversity cap.} At most 3 pairs per Wikipedia article.
\item \emph{Strict A-span admission.} An A prompt is kept only if at least one official SQuAD answer span survives in the stored context with \emph{literal character alignment}: both \texttt{answer\_start + len(answer\_text)} $\leq$ \texttt{len(stored\_context)}, and \texttt{stored\_context[answer\_start : answer\_start + len(answer\_text)] == answer\_text}. Substring matching is not sufficient, because the same string may recur elsewhere in the paragraph.
\item \emph{U-side rule.} U questions come from \texttt{is\_impossible=True} items and have no official answer span by design.
\end{enumerate}

\textbf{Reproducibility.} The shipped file is reproduced exactly by \texttt{src/data/prepare\_squad2.py} with \texttt{split=train}, \texttt{n\_pairs=800}, \texttt{seed=42}.

\textbf{Audit status (2026-04-12).} 1600 rows, 800 complete matched pairs, zero duplicate ids, zero duplicate prompts, zero incomplete pairs; all 1600 prompts uniquely map back to a SQuAD~2.0 train source item, and all 800 A prompts pass literal span-alignment. Earlier versions used a weaker truncation/alignment check that could admit false positives when the answer string recurred elsewhere in the truncated context; the current construction fixes this.

\subsection{falseqa}
\label{app:data-falseqa}

\textbf{Role.} \texttt{falseqa} is used as a \emph{boundary-test dataset} for false-premise QA in the zero-shot transfer evaluation (\S\ref{sec:char-transfer}), \emph{not} as a core structural benchmark.

\textbf{Scale and source.} 1374 prompts (687 A, 687 U), sourced from the public FalseQA dataset.

\textbf{Cleaning procedure.} \texttt{src/data/clean\_falseqa.py} applies the following minimal normalizations while preserving row count, A/U balance, and id structure:
\begin{enumerate}[leftmargin=*,itemsep=2pt,topsep=2pt]
\item \emph{Cross-label duplicate rewrite.} One cross-label exact-duplicate prompt (\texttt{fqa\_0294u} and \texttt{fqa\_0981a} shared the prompt \texttt{``How to light and put out the fire at the same time?''}) is resolved by minimally rewriting the A-side prompt. The script explicitly documents this as a \textbf{local dedup rewrite} rather than a source-faithful restoration of the original FalseQA wording.
\item \emph{Unicode normalization.} Smart quotes (U+2018/2019/201C/201D) are mapped to their ASCII equivalents.
\item \emph{Whitespace collapse.} Repeated spaces are collapsed to single spaces.
\item \emph{Terminal-punctuation repair.} Question-word-initial prompts lacking any terminal punctuation receive \texttt{?}; question-word-initial prompts ending in \texttt{.} (with no mid-sentence \texttt{?}) are rewritten to end in \texttt{?}; imperatives (\texttt{List}/\texttt{Name}/\texttt{Give}) keep or receive \texttt{.}.
\item \emph{Capitalization.} Lowercase \texttt{``If i''} $\to$ \texttt{``If I''} (2 affected rows).
\end{enumerate}

\textbf{Caveat.} After cleaning, token-level length AUC is $0.526$ (Mistral tokenizer, $U = 1$), and the analysis file carries one explicitly documented local dedup rewrite rather than a source-faithful restoration. Because upstream FalseQA has no explicit license, the public artifact does not redistribute this file and instead provides a fetch-and-clean script. Accordingly, \texttt{falseqa} results in this paper are framed as \emph{scope-defining evidence} for false-premise behavior, not as a clean external benchmark.

\subsection{Model release sources}
\label{app:model-sources}

This subsection records canonical citations for every open-weight checkpoint and external dataset used in the paper, concentrated here so that the main body remains untouched by model-release citations.

\textit{Main-grid checkpoints (\S\ref{sec:findings-detection}, \S\ref{sec:findings-ortho}, \S\ref{sec:findings-causal}).} The 11-model main grid is composed of: SmolLM2-1.7B \citep{allal2025smollm2}, Phi-4-mini-3.8B \citep{abdin2025phi4mini}, Mistral-7B-Instruct \citep{jiang2023mistral} and Mistral-Small-24B-Instruct-2501 \citep{mistralai2025small24b}, Gemma-3-4B-it \citep{team2025gemma3}, Llama-3.1-8B-Instruct and Llama-3.3-70B-Instruct (both from the Llama~3 family, \citealp{grattafiori2024llama3}), Qwen3-8B / Qwen3-14B / Qwen3-32B from the Qwen3 family \citep{yang2025qwen3}, and OLMo-2-13B \citep{olmo2024olmo2}.

\textit{Auxiliary checkpoints (\S\ref{sec:findings-pretrain} pretrain comparison and archival appendix tables).} The 6-pair base/instruct comparison in \S\ref{sec:findings-pretrain} adds Qwen2.5-7B/14B/32B-Instruct and their matching base releases from the Qwen2.5 family \citep{yang2024qwen25}, Qwen3-8B-Base and Qwen3-14B-Base \citep{yang2025qwen3}, and Llama-3.1-70B-Base \citep{grattafiori2024llama3}. The legacy energy-decomposition control and detection-heatmap appendix tables (Appendix~\ref{app:energy}, Appendix~\ref{app:per-category-auc}) additionally include Gemma-2 \citep{team2024gemma2}; the ``Phi3'' column in those archival per-category tables is a shorthand label for the Phi-3 release used in the earlier 8-base-cell control and is not part of the 11-model main grid.

\textit{Datasets.} The \texttt{fact800} split is built from SQuAD~2.0 \citep{rajpurkar2018squad2}; \texttt{falseqa} is the cleaned FalseQA file of \citet{hu2023falseqa} (cleaning procedure in App.~\ref{app:data-falseqa}); AbstentionBench-GSM8K (the GSM8K subset of AbstentionBench; \citealp{feng2025abstentionbench,cobbe2021gsm8k}) is the natural-distribution epistemic-style transfer benchmark used in the transfer and length-control analyses of \S\ref{sec:char}.

\section{Additional Detection and Projection Figures}
\label{app:detection-figs}

This appendix collects the detection-side and projection-ablation figures referenced in \S\ref{sec:findings-detection} but not reproduced in the main body for space.

\begin{figure}[h]
  \centering
  \includegraphics[width=\columnwidth]{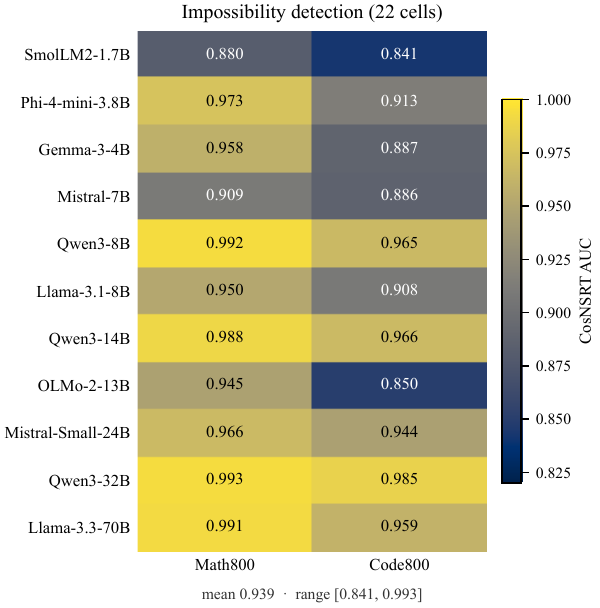}
  \caption{\textbf{CosNSRT global AUC on the 22-cell 11-model main grid (math800 $+$ code800).} Mean AUC 0.939, range $[0.841, 0.993]$. Every 7B+ instruct model exceeds 0.90 on math800; the strongest cell is Qwen3-32B / math800 (0.993), and the weakest is SmolLM2-1.7B / code800 (0.841). Values are HO-AU 5-seed averages at each model's matched layer. Cf. \S\ref{sec:findings-detection}.}
  \label{fig:app-detection-heatmap}
\end{figure}

\begin{figure*}[h]
  \centering
  \includegraphics[width=0.92\textwidth]{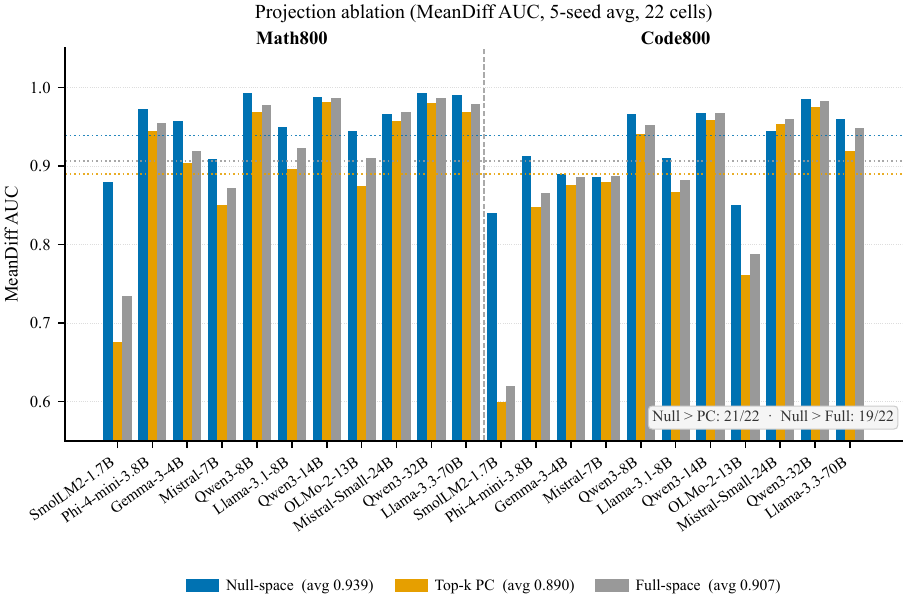}
  \caption{\textbf{MeanDiff AUC under three subspace choices (Null, Top-$k$ PC, Full) on the 22-cell 11-model main grid, 5-seed averaged.} Null-space averages 0.939, Top-$k$ PC 0.890, Full-space 0.907; Null $>$ PC in 21/22 cells, Null $>$ Full in 19/22 cells. The single Null-$\le$-PC cell is Mistral-Small-24B / code800 (gap $-0.009$). Cf. \S\ref{sec:findings-detection}.}
  \label{fig:app-nullspace-ablation}
\end{figure*}

\section{SVM Full-Space Caveat}
\label{app:svm}

\S\ref{sec:findings-detection} attributes the $P \gg w > \varphi$ ordering in the GSRS ablation to the subspace choice: projecting the answerable ($A$) class out of the residual stream raises MeanDiff AUC by $+22.5$pp over a full-space MeanDiff baseline on the 16-cell detection grid, with Null beating Full in 12/16 cells. The framing we give that number in the main text is an \textit{accessibility} result for a low-capacity reader (a one-dimensional cosine probe), not a statement that the impossibility signal exists only in the A-null subspace. This appendix reports the direct evidence for that framing. On the 4 representative (model, dataset) cells for which we have matched SVM-in-null vs.\ SVM-in-full runs (Mistral-7B and Qwen-14B on math800 and code800), a full-space linear SVM matches or exceeds its null-space counterpart in 3 of 4 cells, even though the MeanDiff ordering on those same 4 cells tilts the other way (MeanDiff Null $>$ Full in 3/4).

\begin{table*}[t]
\centering
\small
\setlength{\tabcolsep}{5pt}
\begin{tabular}{@{}llrrrrrr@{}}
\toprule
 & & \multicolumn{3}{c}{MeanDiff AUC} & \multicolumn{3}{c}{SVM AUC} \\
\cmidrule(lr){3-5} \cmidrule(l){6-8}
Model & Dataset & Null & PC & Full & Null & PC & Full \\
\midrule
Mistral-7B & math800 & \textbf{0.907} & 0.851 & 0.872 & 0.982 & 0.951 & \textbf{0.990} \\
Mistral-7B & code800 & \textbf{0.888} & 0.880 & 0.887 & 0.944 & 0.932 & \textbf{0.955} \\
Qwen-14B & math800 & \textbf{0.987} & 0.971 & 0.974 & \textbf{0.996} & 0.986 & 0.996 \\
Qwen-14B & code800 & 0.962 & 0.980 & \textbf{0.982} & 0.990 & 0.976 & \textbf{0.991} \\
\midrule
\multicolumn{2}{l}{Mean} & 0.936 & 0.921 & 0.929 & 0.978 & 0.961 & 0.983 \\
\multicolumn{2}{l}{Null best of 3} & \multicolumn{3}{c}{3/4} & \multicolumn{3}{c}{1/4} \\
\bottomrule
\end{tabular}
\caption{MeanDiff versus SVM AUC under three subspace choices on 4 representative cells. MeanDiff inherits the $P$-dominated ordering of \S\ref{sec:findings-detection} (Null best in 3/4 cells); SVM does not (Full best in 3/4 cells; Qwen-14B/math800 is tied at $0.9963$ Null vs.\ $0.9962$ Full). Bold entries mark the best subspace per row within each classifier family. HO-AU 5-seed averaged.}
\label{tab:svm-caveat}
\end{table*}

Two observations follow. First, the $+22.5$pp $P$-gain reported in \S\ref{sec:findings-detection} is a statement about what a \textit{one-dimensional MeanDiff probe read under cosine similarity} can recover when we change its subspace; it is not a statement about signal existence. On these 4 cells a full-space linear SVM, still a linear classifier but with $D$ free parameters rather than a single direction, closes the gap and on 3 of 4 cells slightly surpasses its A-null-only counterpart. Second, the signs of the MeanDiff and SVM Null-vs-Full comparisons disagree on 3 of 4 cells (Mistral math and code, Qwen-14B code), the pattern one would expect if A-PC variance is a structured nuisance for a low-capacity reader but becomes navigable to a richer classifier that can combine an A-PC component with an A-null component in a weighted sum the cosine-to-a-single-direction probe cannot. We therefore read A-null projection as a method that \textit{makes the signal accessible to simple probes} rather than as a residence claim for impossibility. This framing is compatible with the GSRS ordering $P \gg w > \varphi$ in \S\ref{sec:findings-detection}, which is internal to the MeanDiff family, and is the reason the main-text CosNSRT headline is framed as an accessibility result.

\begin{figure}[h]
  \centering
  \includegraphics[width=0.85\columnwidth]{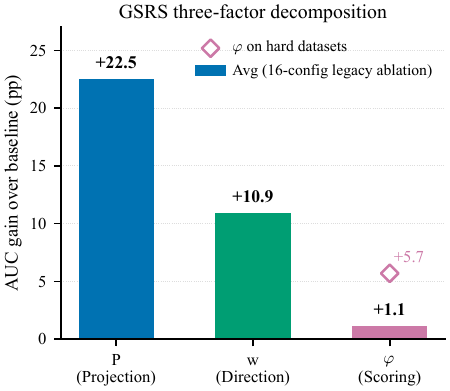}
  \caption{\textbf{Earlier-grid 16-cell GSRS ablation (8 instruct models $\times$ 2 datasets), probe-accessibility evidence behind the $P \gg w > \varphi$ ordering.} Per-factor AUC gain over baseline when only one GSRS factor is swapped onto the best configuration: Projection contributes $+22.5$pp, Direction $+10.9$pp, Scoring $+1.1$pp on average and $+5.7$pp on the hard tail. The figure is a snapshot from the earlier-grid ablation that motivated the A-null subspace choice; the cell list is intentionally not extended to the 11-model main grid. Cf. \S\ref{sec:findings-detection}.}
  \label{fig:app-gsrs-ablation}
\end{figure}

\section{Layer-Emergence Curves}
\label{app:layers}

The main text fixes layer $L$ once per model from a held-out slice (\S\ref{sec:method-hoau}) and reuses that layer across detection, orthogonality, and intervention in most configurations (with one documented exception: Mistral-Small-24B$\times$code800 uses L28 for orthogonality and L20 for steering; see \S\ref{sec:method-hoau}). This appendix reports the descriptive layer-resolved probe that motivated those layer choices. The experiment is a \textit{legacy} 3-model, math800-focused pilot run under an earlier version of the pipeline (V40, April 2026), and we include it only to document the qualitative shape from which the main-text layer selections were made, not as a headline result. It is not a full 8-model sweep, and none of the conclusions in \S\ref{sec:findings} or \S\ref{sec:char} depend on it.

\textit{Protocol (legacy).} On Mistral-7B, Llama-3.1-8B, and Qwen-7B we sweep layers $\ell = 1, 2, \ldots, L_{\max}$ on math800 under two probes: (i) an early-exit linear probe on last-token hidden states (AUC on a held-out split), and (ii) an A-null subspace probe that first fits PCA on train-$A$ and then runs a cross-validated linear probe on the projected residual. Both probes read from the same hidden states; they differ in whether A-subspace variance is projected out before classification. The sweep was densest on the focus layers $\{2, 3, 4, 5, 6, 8\}$ and the final layer, with additional mid-depth probe points at per-model middle layers where the V40 pipeline recorded its peak.

\textit{Shape.} Within the 3-model legacy sweep the per-layer AUC traces an \textit{inverted-U}: the impossibility signal builds up across early and middle layers, peaks at a model-specific middle layer, and either saturates or modestly declines toward the final layer. On Qwen-7B the decline is explicit: the A-null subspace-probe AUC reaches $0.963$ at L17--19 and falls back to $0.845$ at the final layer, while Mistral-7B and Llama-3.1-8B remain close to their middle-layer peaks at the final layer. Table~\ref{tab:layer-peaks} reports the math800 and code800 peak layer and its AUC for each of the three models, as recorded by the V40 middle-depth scan from which the main-text layer choices were derived.

\begin{table}[h]
\centering
\footnotesize
\setlength{\tabcolsep}{2pt}
\resizebox{\columnwidth}{!}{%
\begin{tabular}{@{}lll@{}}
\toprule
Model & math800 peak L (AUC) & code800 peak L (AUC) \\
\midrule
Mistral-7B   & L16 (0.932)     & L15/16 (0.923) \\
Llama-3.1-8B & L13/17 (0.956) & L14 (0.938) \\
Qwen-7B      & L17/19 (0.963) & L18 (0.931) \\
\bottomrule
\end{tabular}%
}
\caption{Legacy middle-layer peaks on the 3-model descriptive probe (math800 and code800, April 2026 V40 pipeline). The main-text layers used throughout \S\ref{sec:findings} (Mistral-7B L15, Llama-3.1-8B L15, Qwen-7B L18) are within one or two layers of these peaks.}
\label{tab:layer-peaks}
\end{table}

Heterogeneity across categories is visible in the legacy logs but not resolved quantitatively: simple symbolic impossibility categories (e.g.\ $\div 0$, $\sqrt{-1}$ over $\mathbb{R}$) tend to reach peak earlier in the network than categories that require multi-step verification (e.g.\ matrix singularity), consistent with the per-category AUC heterogeneity documented in Appendix~\ref{app:per-category-auc}. The main-text layer choice for the 8-model grid (\S\ref{sec:findings-detection}) was informed by these 3-model curves together with per-model exploratory runs: we pick a layer near the middle-layer peak rather than the final layer, a choice the inverted-U shape above justifies qualitatively. We flag this as a heuristic layer-choice procedure and do not re-derive it at the 8-model scale.

\section{Full Per-Category AUC Tables}
\label{app:per-category-auc}

Tables~\ref{tab:per-cat-math} and~\ref{tab:per-cat-code} report the per-category CosNSRT AUC breakdown referenced in \S\ref{sec:char-form} for all 8 instruct models on the 16 math800 and 8 code800 categories. Values are HO-AU 5-seed averaged at each model's matched layer; category identifiers refer to the data-construction scheme detailed in Appendix~\ref{app:data}.

\begin{table*}[t]
\centering
\small
\setlength{\tabcolsep}{5pt}
\begin{tabular}{@{}lcccccccc@{}}
\toprule
\textbf{math800} & SmolLM2 & Gemma2 & Phi3 & Mistral-7B & Qwen-7B & Llama-8B & Qwen-14B & Mistral-Sm. \\
\midrule
\texttt{mdiv}   & 0.994 & 1.000 & 0.995 & 0.997 & 0.970 & 0.994 & 0.999 & 0.986 \\
\texttt{mdive}  & 1.000 & 1.000 & 1.000 & 1.000 & 1.000 & 1.000 & 1.000 & 1.000 \\
\texttt{meven}  & 0.997 & 0.999 & 1.000 & 1.000 & 1.000 & 1.000 & 1.000 & 1.000 \\
\texttt{mfact}  & 0.849 & 0.987 & 0.997 & 0.957 & 1.000 & 0.990 & 1.000 & 1.000 \\
\texttt{mgcd}   & 1.000 & 1.000 & 1.000 & 1.000 & 1.000 & 1.000 & 1.000 & 1.000 \\
\texttt{minv}   & 0.746 & 0.879 & 0.970 & 0.758 & 0.963 & 0.893 & 0.990 & 0.929 \\
\texttt{minva}  & 0.764 & 0.897 & 0.990 & 0.948 & 0.993 & 0.991 & 0.999 & 0.999 \\
\texttt{mlog}   & 0.960 & 0.976 & 0.999 & 0.968 & 0.989 & 0.930 & 1.000 & 1.000 \\
\texttt{mmod}   & 0.994 & 0.997 & 1.000 & 1.000 & 1.000 & 0.999 & 1.000 & 1.000 \\
\texttt{mneg}   & 0.962 & 0.933 & 1.000 & 0.876 & 1.000 & 0.996 & 1.000 & 1.000 \\
\texttt{mno\_r} & 0.761 & 1.000 & 1.000 & 0.991 & 0.888 & 1.000 & 1.000 & 1.000 \\
\texttt{mnone}  & 0.980 & 0.934 & 1.000 & 0.987 & 0.996 & 0.995 & 1.000 & 0.996 \\
\texttt{msing}  & 0.616 & 0.552 & 0.923 & 0.554 & 0.839 & 0.765 & 0.958 & 0.919 \\
\texttt{msqrt}  & 0.999 & 1.000 & 1.000 & 1.000 & 1.000 & 1.000 & 1.000 & 1.000 \\
\texttt{mtrig}  & 0.854 & 0.679 & 0.870 & 0.792 & 0.699 & 0.746 & 0.852 & 0.674 \\
\texttt{muder}  & 0.985 & 1.000 & 1.000 & 0.992 & 0.977 & 0.999 & 1.000 & 1.000 \\
\midrule
\textbf{Global} & 0.880 & 0.915 & 0.981 & 0.908 & 0.953 & 0.952 & 0.987 & 0.965 \\
\bottomrule
\end{tabular}
\caption{Per-category CosNSRT AUC on math800 (16 categories $\times$ 8 models). Global AUC is the HO-AU 5-seed mean over the full dataset at the model's matched layer.}
\label{tab:per-cat-math}
\end{table*}

\begin{table*}[t]
\centering
\small
\setlength{\tabcolsep}{5pt}
\begin{tabular}{@{}lcccccccc@{}}
\toprule
\textbf{code800} & SmolLM2 & Gemma2 & Phi3 & Mistral-7B & Qwen-7B & Llama-8B & Qwen-14B & Mistral-Sm. \\
\midrule
\texttt{cattr} & 0.799 & 0.625 & 0.777 & 0.670 & 0.864 & 0.761 & 0.892 & 0.799 \\
\texttt{cinde} & 0.584 & 0.714 & 0.828 & 0.840 & 0.897 & 0.861 & 0.978 & 0.924 \\
\texttt{cinfi} & 0.981 & 0.990 & 0.993 & 0.988 & 0.995 & 0.969 & 0.960 & 0.995 \\
\texttt{cmath} & 0.825 & 0.935 & 0.974 & 0.925 & 0.951 & 0.980 & 1.000 & 0.995 \\
\texttt{cname} & 0.996 & 0.991 & 1.000 & 0.989 & 0.993 & 0.990 & 0.995 & 0.966 \\
\texttt{ctype} & 0.853 & 0.837 & 0.893 & 0.872 & 0.908 & 0.843 & 0.958 & 0.951 \\
\texttt{cvalu} & 0.651 & 0.756 & 0.830 & 0.795 & 0.849 & 0.860 & 0.916 & 0.891 \\
\texttt{czero} & 0.992 & 0.995 & 0.997 & 0.997 & 0.976 & 0.997 & 1.000 & 1.000 \\
\midrule
\textbf{Global} & 0.840 & 0.866 & 0.915 & 0.884 & 0.928 & 0.908 & 0.961 & 0.944 \\
\bottomrule
\end{tabular}
\caption{Per-category CosNSRT AUC on code800 (8 categories $\times$ 8 models). Global AUC is the HO-AU 5-seed mean over the full dataset at the model's matched layer.}
\label{tab:per-cat-code}
\end{table*}

\section{Steering Full \texorpdfstring{$\alpha$}{alpha} Sweep and max-\texorpdfstring{$\Delta$}{Delta} Envelope}
\label{app:steering}

\textbf{Note (legacy v1 archival).} The 23-cell breadth analysis below is v1 archival. The current \S\ref{sec:findings-causal} headline uses the v2-deterministic 48-cell steering breadth (16 models $\times$ 3 datasets) documented in Appendix~\ref{app:protocol-audit}. The table and per-cell numbers below are retained as a reproducibility artifact for the v1 protocol, not as support for the current headline. The v1 protocol reported steering as a signal-minus-random refusal-rate-on-U gain at best $\alpha$, with one cell (Mistral-Small-24B$\times$fact800) absent at the time of recording. For each cell the table shows two pairs of numbers: (i) at the \textit{best $\alpha$}, the signal-minus-random gains on the correct behavior (refusal on U; $\Delta_{\mathrm{refU}}$) and on the selectivity cost (wrong refusal on A; $\Delta_{\mathrm{wrongA}}$); and (ii) the $\alpha$ at which $\Delta_{\mathrm{refU}}$ is maximized anywhere across $\{0, 5, 10, 20, 30, 40\} \cdot \mathrm{proj\_std}$, and the resulting max-$\Delta$ envelope. The max-$\alpha$ columns document the full v1 causal-capacity envelope of the handle on this archival grid; readers should consult Appendix~\ref{app:protocol-audit} and \S\ref{sec:findings-causal} for the current v2 16-model 48-cell breadth.

\begin{table*}[t]
\centering
\small
\setlength{\tabcolsep}{4pt}
\begin{tabular}{@{}llrrrrrrr@{}}
\toprule
 & & & \multicolumn{3}{c}{at best $\alpha$} & \multicolumn{3}{c}{at max-$\Delta$ $\alpha$} \\
\cmidrule(lr){4-6} \cmidrule(l){7-9}
Model & Dataset & L & best $\alpha$ & $\Delta_{\mathrm{refU}}$ & $\Delta_{\mathrm{wrongA}}$ & max $\alpha$ & $\Delta_{\mathrm{refU}}$ & $\Delta_{\mathrm{wrongA}}$ \\
\midrule
SmolLM2      & math800 & 11 & 5  & $+0.00$ & $-0.01$ & 0  & $+0.00$ & $+0.00$ \\
Gemma2       & math800 & 16 & 20 & $+0.43$ & $+0.19$ & 20 & $+0.43$ & $+0.19$ \\
Phi3         & math800 & 15 & 5  & $+0.21$ & $+0.10$ & 30 & $+0.91$ & $+0.85$ \\
Mistral-7B   & math800 & 15 & 5  & $+0.25$ & $+0.02$ & 10 & $+0.45$ & $+0.29$ \\
Qwen-7B      & math800 & 18 & 5  & $+0.02$ & $+0.02$ & 20 & $+0.31$ & $+0.57$ \\
Llama-8B     & math800 & 15 & 5  & $+0.26$ & $-0.02$ & 20 & $+0.86$ & $+0.81$ \\
Qwen-14B     & math800 & 34 & 5  & $+0.06$ & $+0.01$ & 30 & $+1.00$ & $+1.00$ \\
Mistral-Sm.  & math800 & 28 & 10 & $+0.11$ & $+0.02$ & 40 & $+0.68$ & $+0.54$ \\
\midrule
SmolLM2      & code800 & 14 & 0  & $+0.00$ & $+0.00$ & 40 & $+0.01$ & $+0.03$ \\
Gemma2       & code800 & 14 & 10 & $+0.17$ & $+0.09$ & 40 & $+0.93$ & $+0.91$ \\
Phi3         & code800 & 16 & 10 & $+0.12$ & $+0.06$ & 20 & $+0.23$ & $+0.18$ \\
Mistral-7B   & code800 & 15 & 20 & $+0.32$ & $+0.16$ & 40 & $+0.57$ & $+0.52$ \\
Qwen-7B      & code800 & 18 & 20 & $+0.00$ & $+0.00$ & 40 & $+0.08$ & $+0.26$ \\
Llama-8B     & code800 & 14 & 5  & $+0.15$ & $+0.06$ & 30 & $+0.83$ & $+0.66$ \\
Qwen-14B     & code800 & 32 & 10 & $+0.02$ & $-0.03$ & 20 & $+0.10$ & $+0.03$ \\
Mistral-Sm.  & code800 & 20 & 10 & $+0.14$ & $+0.01$ & 10 & $+0.14$ & $+0.01$ \\
\midrule
SmolLM2      & fact800 & 11 & 5  & $+0.00$ & $-0.01$ & 0  & $+0.00$ & $+0.00$ \\
Gemma2       & fact800 & 16 & 10 & $+0.26$ & $+0.13$ & 20 & $+0.63$ & $+0.47$ \\
Phi3         & fact800 & 15 & 20 & $+0.17$ & $+0.08$ & 40 & $+0.40$ & $+0.53$ \\
Mistral-7B   & fact800 & 17 & 20 & $+0.14$ & $+0.03$ & 40 & $+0.42$ & $+0.28$ \\
Qwen-7B      & fact800 & 19 & 0  & $+0.00$ & $+0.00$ & 30 & $+0.49$ & $+0.25$ \\
Llama-8B     & fact800 & 15 & 10 & $+0.25$ & $+0.17$ & 30 & $+0.66$ & $+0.66$ \\
Qwen-14B     & fact800 & 34 & 10 & $+0.01$ & $+0.03$ & 20 & $+0.20$ & $+0.27$ \\
\bottomrule
\end{tabular}
\caption{Legacy v1 8-model 23-cell archival summary of the full $\alpha$-sweep (signal-minus-random refusal-rate-on-U gain). $\alpha$ is reported in units of proj\_std (the standard deviation of null-space projections of train hidden states onto $\hat{d}$, \S\ref{sec:method-intervention}); $\Delta_{\mathrm{refU}}$ and $\Delta_{\mathrm{wrongA}}$ are the signal-minus-random differences in $\mathrm{refusal\_rate}_U$ and $\mathrm{wrong\_refusal\_rate}_A$ at that $\alpha$. The best $\alpha$ is the single value per cell that maximizes the cell-level \textit{overall proxy} $(n_{\mathrm{correct\_refusal}_U} + n_{\mathrm{non\_refusal}_A}) / n_{\mathrm{total}}$; the max-$\Delta$ $\alpha$ is the value that maximizes $\Delta_{\mathrm{refU}}$ in isolation. The per-domain best-$\alpha$ column maxima and the max-$\Delta$ envelopes are archival v1 figures only. For the current v2-deterministic 48-cell steering breadth on 16 models, see \S\ref{sec:findings-causal} and Appendix~\ref{app:protocol-audit}.}
\label{tab:steering-sweep}
\end{table*}

\textit{Two reading-level cautions (scoped to the legacy v1 8-model archival table).} First, the max-$\alpha$ envelope is an \textit{existence proof} that $d_{\mathrm{imp}}$ can drive large behavioral changes relative to random on this 8-model 23-cell grid, not an operating recommendation: the $\Delta_{\mathrm{wrongA}}$ column in the max block is of comparable magnitude to $\Delta_{\mathrm{refU}}$ on most cells, which means that at those $\alpha$ values the model is also refusing the answerable class; high $\Delta_{\mathrm{refU}}$ comes with a proportional selectivity cost. The Qwen-14B$\times$math800 case is the extreme form on this archival grid: max-$\alpha = 30$ delivers $\Delta_{\mathrm{refU}} = +1.00$ but also $\Delta_{\mathrm{wrongA}} = +1.00$, i.e., the model refuses \textit{everything} under steering and \textit{nothing} under the random-direction control. Second, within this archival table the per-domain ordering math $>$ code $>$ fact in best-$\alpha$ gains reproduces in the max-$\Delta$ envelope; that ordering is internal to the v1 archival reference. It is \textit{distinct} from, not a substitute for, the current v2 evidence in \S\ref{sec:findings-causal} (anchor-quality math/code with Mistral-7B as bidirectional keystone, fact as structural boundary) and Appendix~\ref{app:protocol-audit}; the two views should not be conflated.

\section{Legacy v1 Keyword-Detector Audit}
\label{app:human}

\textbf{Note (legacy v1 archival).} The 4,500-row keyword-detector audit described in this appendix is v1 archival. The current \S\ref{sec:findings-causal} headline numbers come from a v2 invalidity-aware LLM-assisted rubric documented in Appendix~\ref{app:protocol-audit}, run on the v2 4-anchor grid (Mistral-7B-Instruct, Gemma-3-4B-it, Qwen3-14B, Qwen3-8B), not the v1 3$\times$3 grid that this appendix documents. The keyword-detector audit below is retained as a reproducibility artifact for the v1 protocol, not as support for the current headline.

\textit{Pipeline.} The v1 intervention pipeline detected refusal at generation time with a keyword lexicon: an output was flagged as a refusal if it contained any of a hand-curated list of refusal tokens (e.g.\ ``cannot'', ``undefined'', ``no solution'', domain-specific equivalents for code and math). The v1 \textit{gated flip rate} metric documented in this appendix was computed on this detector. For the legacy audit, an LLM-assisted batch review assigned three-way labels to all 4{,}500 intervention records under a fixed rubric, with the first author reviewing uncertain cases. Each row stores both signal- and random-branch generations, but its retained three-way label pertains to the signal-branch detector decision; the random branch was not separately annotated.

\textit{Result.} Table~\ref{tab:human-verification} reports 747 flip labels, 3{,}749 no-flip labels, and 4 keyword-detector false positives. The recorded detector-disagreement share is $4 / 4{,}500 = 0.09\%$ overall and $4 / 751 \approx 0.53\%$ among detector-positive rows. The 4 errors are concentrated in 2 of the 9 cells (3 in Qwen-14B$\times$fact800; 1 in Mistral-7B$\times$code800) and are absent on every math800 cell.

\begin{table}[h]
\centering
\small
\setlength{\tabcolsep}{3.5pt}
\begin{tabular}{@{}llrrrr@{}}
\toprule
Model & Dataset & Rows & Flip & No flip & FP \\
\midrule
Mistral-7B & math800 & 500 & 150 & 350 & 0 \\
Mistral-7B & code800 & 500 & 64  & 435 & 1 \\
Mistral-7B & fact800 & 500 & 19  & 481 & 0 \\
Qwen-7B    & math800 & 500 & 101 & 399 & 0 \\
Qwen-7B    & code800 & 500 & 31  & 469 & 0 \\
Qwen-7B    & fact800 & 500 & 60  & 440 & 0 \\
Qwen-14B   & math800 & 500 & 213 & 287 & 0 \\
Qwen-14B   & code800 & 500 & 69  & 431 & 0 \\
Qwen-14B   & fact800 & 500 & 40  & 457 & 3 \\
\midrule
\textbf{Total} & & \textbf{4{,}500} & \textbf{747} & \textbf{3{,}749} & \textbf{4} \\
\bottomrule
\end{tabular}
\caption{Legacy v1 LLM-assisted audit of the signal-branch keyword detector across 9 model--dataset cells. ``FP'' counts recorded detector disagreements. Each row totals 500. The disagreement share is $4 / 4{,}500 = 0.09\%$ overall and $4 / 751 \approx 0.53\%$ among detector-positive rows.}
\label{tab:human-verification}
\end{table}

\textit{Implications for the v1 archival result.} All three v1 math800 cells have zero recorded false-positive keyword detections across 1{,}500 audited rows; on this archival v1 3$\times$3 grid the recorded disagreement share was $\approx 0.09\%$ overall / $\approx 0.53\%$ among detector-positive rows. The 4 errors that did occur clustered in v1 fact800 (3 of 4, all on Qwen-14B), consistent with fact800 being an entangled domain on the v1 protocol. The current \S\ref{sec:findings-causal} headline numbers do \textit{not} come from this keyword detector; they are produced by the v2 invalidity-aware classifier on the v2 4-anchor grid (Appendix~\ref{app:protocol-audit}). The v1 audit reported here bounds the v1 archival numbers only, not the current v2 headline.

\section{Negative Results}
\label{app:negatives}
Over the course of pipeline development, we evaluated a systematic family of ``internal mismatch'' hypotheses for structural unanswerability: the hypothesis that A/U could be distinguished by \emph{disagreement} or \emph{split} between different internal components of the forward pass (attention pattern vs.\ gradient, layer trajectories vs.\ output, attention entropy vs.\ representation own-distance, etc.). Table~\ref{tab:negative-hypotheses} lists 12 such hypotheses, evaluated on the early 3-model \texttt{math50} pilot and, where a pilot result was borderline, re-run on the full-scale \texttt{math800} dataset. None reached usable discriminative performance. This body of negative evidence is what motivated the shift to the A-null residual-space direction $d_{\mathrm{imp}}$ used in \S\ref{sec:findings-detection}.

\begin{table*}[t]
\centering
\footnotesize
\setlength{\tabcolsep}{4pt}
\begin{tabular}{@{}r>{\raggedright\arraybackslash}p{4.65cm}>{\raggedright\arraybackslash}p{3.55cm}>{\centering\arraybackslash}p{1.65cm}>{\centering\arraybackslash}p{1.55cm}>{\raggedright\arraybackslash}p{2.85cm}@{}}
\toprule
\textbf{\#} & \textbf{Hypothesis} & \textbf{Signal family} & \textbf{\texttt{math50}} & \textbf{\texttt{math800}} & \textbf{Failure mode} \\
\midrule
1 & Layer-wise $ae{\downarrow}$ but $od{\uparrow}$ mismatch & Attention entropy $\times$ own-distance & 0.56--0.65 & 0.51--0.57 & Chance \\
2 & Per-token attention--gradient divergence & Attention $+$ gradient & 0.615 & --- & Chance \\
3 & Representation trajectory growth (deep$-$shallow $od$) & Own-distance & --- & 0.54--0.55 & Chance \\
4 & Cross-head attention split degree & Attention pattern & --- & --- & Direction reversed (Qwen) \\
5 & Output logit entropy vs.\ $od$ mismatch & Logit entropy & --- & --- & Direction inverted \\
6 & Inter-head attention variance & Attention pattern & --- & --- & Chance \\
7 & U-deviation directional consistency & Projection onto mean U-direction & 0.783 & --- & Weakened linear-probe variant \\
8 & Attn/MLP decomposition coordination & \texttt{sign\_agree}, \texttt{pearson\_r}, \texttt{abs\_diff} & 0.44--0.59 & 0.50--0.55 & Chance \\
9 & Trajectory curvature (consecutive-update cosine) & Layer trajectory & 0.44--0.52 & 0.50--0.57 & Chance \\
10 & Layer-segment $\Delta_{\mathrm{attn}} - \Delta_{\mathrm{mlp}}$ & Attn/MLP decomposition & Weak/\allowbreak{}unstable & --- & Chance \\
11 & Same-layer $ae$--$od$ mismatch (4 formulations) & Attention entropy $\times$ own-distance & 0.51--0.72 & 0.50--0.59 & Chance \\
12 & Token-level geometric attribution ($\partial od / \partial e_t$) & Gradient attribution & 0.63--0.65 & --- & $r{=}0.76$ with $od$ (redundant) \\
\bottomrule
\end{tabular}
\caption{12 internal-mismatch hypotheses ruled out during pipeline development. ``---'' marks cells that were not run at that scale because a cheaper pilot variant had already failed. \texttt{math800} AUC entries are aggregated across 3 models $\times$ 16 categories $\times$ 50 A/U pairs. Column abbreviations: $ae$ = attention entropy, $od$ = representation own-distance.}
\label{tab:negative-hypotheses}
\end{table*}

\textbf{Root-cause analysis.} The uniform failure across the 12 hypotheses has a single underlying explanation. Uncertainty associated with \emph{structural} unanswerability propagates \emph{coherently} through the forward pass: representation, attention, gradient, and output-distribution signals all move in the same direction on a U prompt relative to its matched A prompt, so no consistent internal split between components exists to detect. These mismatch hypotheses were designed to detect component disagreement, such as an attention pattern that does not match a confidently emitted token. Structural impossibility instead leaves a coherent mark along a specific geometric axis of the residual stream, namely the A-null direction $d_{\mathrm{imp}}$ used in the main text. The recognition signal is \emph{concentrated along a subspace}, not distributed across component mismatches, which is why every mismatch-based formulation we tried did not yield discriminative performance and why a simple one-dimensional null-space MeanDiff does.

\textbf{Other negative findings (outside the 12 internal-mismatch hypotheses).} Three additional negative results shaped the final pipeline and are noted here for completeness.

\begin{enumerate}[leftmargin=*,itemsep=2pt,topsep=2pt]
\item \emph{$\tau$ (attractor tightness) as a cross-dataset predictor.} An earlier claim that across-dataset variation in detection AUC tracks $\tau$ with Spearman $\rho = -0.950$ did not survive replication on the current pipeline ($\rho = -0.147$, $p = 0.65$, $n = 12$). Within-dataset NS-SNR (null-space signal-to-noise ratio) vs.\ detection AUC ($\rho = 0.76$) replaces it as the operative geometric summary.
\item \emph{Semantic entropy as an impossibility baseline.} Output-distribution semantic entropy \citep{farquhar2024detecting} on \texttt{llama}/\texttt{math50} (100 prompts, 10 sampled generations each, non-strict entailment) yields $\mathrm{AUC} = 0.625$ (cluster-assignment entropy $= 0.638$), while the representation-level geometry probe on the \emph{same samples} yields $\mathrm{AUC} = 0.841$; the null-space direction $d_{\mathrm{imp}}$ used in the main text has mean $\mathrm{AUC} = 0.924$ across 8 instruct models $\times$ 2 structural-impossibility domains (\S\ref{sec:findings-detection}). Output-level uncertainty signals therefore carry meaningfully less impossibility information than residual-stream null-space geometry.
\item \emph{Unsupervised discovery of $d_{\mathrm{imp}}$.} Four label-free methods (PCA on A-null residuals, kurtosis maximization, skewness maximization, and FastICA) were applied to the same A-null residuals that CosNSRT consumes. The best unsupervised direction achieves $\mathrm{AUC} = 0.675$ with $\cos(\hat d, d_{\mathrm{imp}}) < 0.48$, substantially below the labeled MeanDiff direction. A/U labels are therefore a necessary input to recover the impossibility axis from the A-null residual stream; the direction is not the dominant unsupervised axis of that subspace.
\end{enumerate}

\section{Energy Decomposition Full Tables}
\label{app:energy}

The main text claims, in the orthogonality analysis (\S\ref{sec:findings-ortho}), that the moderate-to-substantial full-space cosine $\cos_{\mathrm{full},\mathrm{full}}(d_{\mathrm{imp}}^{\mathrm{full}}, d_{\mathrm{ref,safety}}) \in [0.057, 0.781]$ across the 22 main-grid instruct cells is dominated by shared answerable-structure variance rather than by shared impossibility signal: averaged over the 22 instruct cells, the PC-PC component accounts for mean $0.813$ of the magnitude of $\cos_{\mathrm{full},\mathrm{full}}$ (range $[0.607, 0.983]$). Table~\ref{tab:energy-instruct} reports the per-cell attribution behind that mean. Table~\ref{tab:energy-base} reports an auxiliary base-model decomposition on an earlier-grid 14-cell instruct/base set; this is auxiliary base-model decomposition, not part of the 22-cell instruct main-grid control, and is not the source of the \S\ref{sec:findings-pretrain} 6-pair pretraining-origin headline. We decompose $\cos_{\mathrm{full},\mathrm{full}} = \mathrm{PC\text{-}PC} + \mathrm{null\text{-}null}$ (the two cross terms vanish because $V_k^{\top} V_k$ and $I - V_k^{\top} V_k$ are orthogonal projectors), compute $\mathrm{null\text{-}null} = \sqrt{E_{\mathrm{null}}(d_{\mathrm{imp}}^{\mathrm{full}})} \cdot \cos_{\mathrm{matched},\mathrm{full}}$ from the saved rep files, and set $\mathrm{PC\text{-}PC}$ by subtraction; ``PC-PC share'' and ``null-null share'' are the magnitude shares $|\mathrm{PC\text{-}PC}|/(|\mathrm{PC\text{-}PC}| + |\mathrm{null\text{-}null}|)$ and its complement. Both shares sum to 1 by construction; on all 22 main-grid instruct cells, $\mathrm{PC\text{-}PC}$ and $\mathrm{null\text{-}null}$ have the same sign as $\cos_{\mathrm{full},\mathrm{full}}$, so magnitude share equals signed share.

\begin{table}[t]
\centering
\footnotesize
\setlength{\tabcolsep}{2pt}
\resizebox{\columnwidth}{!}{%
\begin{tabular}{@{}llrrrr@{}}
\toprule
Model & Dataset & $\cos_{\mathrm{m},\mathrm{f}}$ & $\cos_{\mathrm{f},\mathrm{f}}$ & PC-PC & null-null \\
\midrule
SmolLM2-1.7B    & math800 & $+0.052$ & $+0.101$ & 0.698 & 0.302 \\
SmolLM2-1.7B    & code800 & $+0.048$ & $+0.154$ & 0.877 & 0.123 \\
Phi-4-mini-3.8B & math800 & $+0.124$ & $+0.252$ & 0.801 & 0.199 \\
Phi-4-mini-3.8B & code800 & $+0.075$ & $+0.118$ & 0.744 & 0.256 \\
Gemma-3-4B      & math800 & $+0.122$ & $+0.531$ & 0.935 & 0.065 \\
Gemma-3-4B      & code800 & $+0.086$ & $+0.781$ & 0.983 & 0.017 \\
Mistral-7B      & math800 & $+0.120$ & $+0.272$ & 0.809 & 0.191 \\
Mistral-7B      & code800 & $+0.090$ & $+0.245$ & 0.897 & 0.103 \\
Qwen3-8B        & math800 & $+0.109$ & $+0.238$ & 0.798 & 0.202 \\
Qwen3-8B        & code800 & $+0.116$ & $+0.278$ & 0.826 & 0.174 \\
Llama-3.1-8B    & math800 & $+0.068$ & $+0.257$ & 0.888 & 0.112 \\
Llama-3.1-8B    & code800 & $+0.072$ & $+0.244$ & 0.880 & 0.120 \\
Qwen3-14B       & math800 & $+0.080$ & $+0.251$ & 0.876 & 0.124 \\
Qwen3-14B       & code800 & $+0.130$ & $+0.234$ & 0.752 & 0.248 \\
OLMo-2-13B      & math800 & $+0.101$ & $+0.197$ & 0.761 & 0.239 \\
OLMo-2-13B      & code800 & $+0.097$ & $+0.104$ & 0.607 & 0.393 \\
Mistral-Sm.-24B & math800 & $+0.065$ & $+0.123$ & 0.808 & 0.192 \\
Mistral-Sm.-24B & code800 & $+0.019$ & $+0.180$ & 0.963 & 0.037 \\
Qwen3-32B       & math800 & $+0.098$ & $+0.240$ & 0.819 & 0.181 \\
Qwen3-32B       & code800 & $+0.110$ & $+0.192$ & 0.719 & 0.281 \\
Llama-3.3-70B   & math800 & $+0.098$ & $+0.239$ & 0.817 & 0.183 \\
Llama-3.3-70B   & code800 & $+0.037$ & $+0.056$ & 0.617 & 0.383 \\
\midrule
\multicolumn{2}{l}{\textbf{Mean}} & $+0.087$ & $+0.240$ & \textbf{0.813} & 0.187 \\
\bottomrule
\end{tabular}%
}
\caption{Energy decomposition of $\cos_{\mathrm{full},\mathrm{full}}$ on the 22 main-grid instruct cells (11 models $\times$ \{math800, code800\}). $\cos_{\mathrm{m},\mathrm{f}}$ is the matched-layer cosine used in the \S\ref{sec:findings-ortho} headline (with $d_{\mathrm{imp}}$ restricted to A-null and $d_{\mathrm{ref,safety}}$ in full space); $\cos_{\mathrm{f},\mathrm{f}}$ is the full-space-both cosine. PC-PC and null-null are magnitude shares, summing to 1. Mean PC-PC share across the 22 cells is $0.813$, range $[0.607, 0.983]$.}
\label{tab:energy-instruct}
\end{table}

\begin{table}[t]
\centering
\footnotesize
\setlength{\tabcolsep}{2pt}
\begin{tabular}{@{}llrrrr@{}}
\toprule
Model (base) & Dataset & $\cos_{\mathrm{m},\mathrm{f}}$ & $\cos_{\mathrm{f},\mathrm{f}}$ & PC-PC & null-null \\
\midrule
Mistral-7B   & math800 & $+0.070$ & $+0.180$ & 0.844 & 0.156 \\
Mistral-7B   & code800 & $+0.063$ & $+0.145$ & 0.862 & 0.138 \\
Llama-8B     & math800 & $+0.060$ & $+0.190$ & 0.849 & 0.151 \\
Llama-8B     & code800 & $+0.073$ & $+0.131$ & 0.809 & 0.191 \\
Qwen-7B      & math800 & $+0.122$ & $+0.305$ & 0.812 & 0.188 \\
Qwen-7B      & code800 & $+0.040$ & $+0.324$ & 0.971 & 0.029 \\
SmolLM2      & math800 & $+0.060$ & $+0.121$ & 0.715 & 0.285 \\
SmolLM2      & code800 & $+0.059$ & $+0.145$ & 0.835 & 0.165 \\
Gemma2       & math800 & $+0.045$ & $+0.080$ & 0.740 & 0.260 \\
Gemma2       & code800 & $+0.013$ & $+0.078$ & 0.948 & 0.052 \\
Qwen-14B     & math800 & $+0.056$ & $+0.209$ & 0.887 & 0.113 \\
Qwen-14B     & code800 & $+0.081$ & $+0.212$ & 0.864 & 0.136 \\
Mistral-Sm.  & math800 & $+0.066$ & $+0.067$ & 0.619 & 0.381 \\
Mistral-Sm.  & code800 & $-0.004$ & $+0.047$ & 0.980 & 0.020 \\
\midrule
\multicolumn{2}{l}{\textbf{Mean}} & $+0.058$ & $+0.160$ & \textbf{0.838} & 0.162 \\
\bottomrule
\end{tabular}
\caption{\textit{Auxiliary base-model decomposition, not part of the 22-cell instruct main-grid control of Table~\ref{tab:energy-instruct}, and not the source of the \S\ref{sec:findings-pretrain} 6-pair pretraining-origin headline.} Energy decomposition on a 14-cell base set under an earlier model grid (7 base models $\times$ 2 datasets; Phi3 has no matching base checkpoint). Reported here only as a higher-resolution view that the same shared-answerable-structure decomposition pattern reproduces under base post-training; mean PC-PC share is $0.838$ on this auxiliary set.}
\label{tab:energy-base}
\end{table}

\textit{What the tables show, and what they do not.} Across the 22 main-grid instruct cells, PC-PC share is $\geq 0.70$ on 19/22 cells (the three below-threshold cells are SmolLM2-1.7B/math800 at $0.698$, OLMo-2-13B/code800 at $0.607$, and Llama-3.3-70B/code800 at $0.617$, all small-$\cos_{\mathrm{full},\mathrm{full}}$ cells where PC-PC and null-null are individually small and the share fraction is correspondingly more sensitive); null-null share never exceeds $0.40$ on any cell. Equivalently, the full-space cosine $\cos_{\mathrm{full},\mathrm{full}}$, which on first reading of \S\ref{sec:findings-ortho} might look like moderate-to-substantial alignment between $d_{\mathrm{imp}}$ and $d_{\mathrm{ref,safety}}$, is in every cell reassembled almost entirely from $d_{\mathrm{imp}}^{\mathrm{full}}$'s A-PC component and $d_{\mathrm{ref,safety}}$'s A-PC component, both of which encode shared answerable-structure variance (topic, surface form, syntactic scaffolding). The null-null component, where a genuine shared-impossibility signal would live, contributes a mean magnitude share of $0.187$ across the 22 main-grid instruct cells. This is the quantitative object behind the \S\ref{sec:findings-ortho} statement ``in the subspace where the impossibility signal actually lives, A-null, the two directions remain near-orthogonal.'' The tables do not, by themselves, exclude that $d_{\mathrm{ref,safety}}$ picks up A/U predictive power from the shared A-PC subspace; indeed \S\ref{sec:findings-ortho} reports $d_{\mathrm{ref,safety}}$'s A/U AUC on math800 in range $[0.600, 0.962]$, consistent with the shared A-PC overlap reported here.

\section{Protocol Refinement Audit (v1 vs.\ v2)}
\label{app:protocol-audit}

The intervention and steering analyses in \S\ref{sec:findings-causal} share an invalidity-aware verification rubric (with mixed-output and degenerate-output guards) but apply it differently. The intervention grid (4 anchors $\times$ 3 datasets) uses candidate labels assigned to all records under a fixed written rubric through an LLM-assisted batch review, supported by deterministic domain-specific labeling utilities. A second pass covered flagged subsets, and the first author reviewed uncertain cases; the first author did not independently review every row. In the nine non-Qwen3-8B cells, the aggregate results apply candidate labels plus provisional second-pass audit fills; the three Qwen3-8B cells apply candidate labels without second-pass overrides. Accordingly, the v2 grid is LLM-assisted rather than fully human-adjudicated. The v1 version of the same grid used a keyword-based refusal detector with the LLM-assisted batch audit documented in Appendix~\ref{app:human}, and v2 refines that procedure under the new rubric. The steering breadth sweep (16 models $\times$ 3 datasets) uses a deterministic regex-based proxy aligned with the v2 intervention rubric, re-aggregating row-level steering outputs without human relabeling; the v1 steering aggregation used a lexical refusal-keyword detector. Neither protocol re-ran model generation; both versions use the same row-level outputs. This appendix documents the v1 $\to$ v2 refinement on each protocol, the cells where it changes the headline, and the v2-only diagnostics that motivated the refinement.

\subsection{What v2 Adds}
\label{app:protocol-audit-deltas}

\begin{itemize}
\item \textbf{Invalidity-aware vocabulary} (vs.\ v1 lexical refusal list). The v2 classifier accepts domain-specific invalidity recognition as abstention: ``undefined'' / ``no real solution'' / ``diverges'' on math; ``raises TypeError / ValueError / ZeroDivisionError / IndexError'' / ``unsupported operand'' on code; ``not stated in the passage'' / ``passage does not provide'' on fact. The legacy keyword list missed most exception-named code abstentions and some fact-passage abstention forms.
\item \textbf{Mixed-output guard}. If a generation gives a concrete normal answer in the first $\sim$120 characters and then appends an irrelevant invalidity caveat (e.g.\ ``The result is $20$. However, indexing out of range raises \texttt{IndexError}.''), the row is labeled not-abstention. The legacy keyword detector classified such rows as abstention based on the appended caveat.
\item \textbf{Degenerate-aware computation}. Branches whose generation collapses into token-soup, sentence-level repetition, or recursive nested syntax that never terminates are kept in the gated denominator but force flip false; clean baselines that themselves degenerate are excluded from the cell. The legacy aggregation silently included degenerate branches with whatever keyword decision the lexical detector emitted.
\item \textbf{Empty-gate handling}. Cells whose gated denominator is zero under a criterion are reported as not-measurable (N/A), not as zero. The legacy aggregation reported $0.0$, conflating ``no clean baseline to measure on'' with ``intervention had no effect''.
\item \textbf{Mixed candidate labeling with second-pass review} (intervention grid only; the steering breadth sweep uses a deterministic regex-based proxy with no relabeling). Candidate labels are assigned to all records under a fixed written rubric through an LLM-assisted batch review, supported by deterministic domain-specific labeling utilities and followed by schema and gate validation. The audit set contains candidate flips, uncertain and degenerate rows, clean baselines, and a $10\%$ stratified sample of non-flips; the first author reviewed uncertain cases but did not independently review every row. The nine non-Qwen3-8B result files apply provisional LLM-assisted second-pass audit fills to this subset; the three Qwen3-8B result files apply candidate-label passthrough. The second pass exposes both directions of mixed-output misclassification observed during calibration ($27\%$ over-credit on the Mistral code provisional candidate-flip subset; $2/300$ over-strict on the Gemma-3-4B fact checked rows).
\end{itemize}

\subsection{v1 $\to$ v2 Headline Shifts}
\label{app:protocol-audit-shifts}

Counted strictly per slot across the $24$ (anchor, dataset, direction) cells in the 4-anchor intervention grid, $17$ decrease relative to v1 best-$\alpha$ gated $\Delta$G, $5$ increase, $1$ is unchanged, and $1$ becomes unmeasurable. The earlier 18-down / 4-up / 2-flat summary uses a $6$pp flatness tolerance and groups the newly unmeasurable Mistral fact U$\to$A slot with the downshifts. The dominant causes (cross-tabulated against each cell in the supplementary factsheet) are:

\begin{itemize}
\item \textbf{Mixed-output false-positive catch} (largest single cause): the primary cause in $9$ of the $10$ A$\to$U downshifts. In these cells, the v1 keyword detector counted ``concrete answer + appended invalidity caveat'' rows as successful flips; the v2 mixed-output guard correctly rejects them. Mistral code A$\to$U is the cleanest example: re-audit of $37$ candidate flip rows under the strict mixed-output rule overrode $10$ ($27\%$), and the post-override gated $\Delta$G ($+12.5$pp at $\alpha{=}20$ and $+35.4$pp at $\alpha{=}40$) matches the prior-keyword $\Delta$G on the same cell exactly, evidence that the v1 numbers on this cell were correct on average but happened to be inflated on the particular candidate-flip subset that the mixed-output guard targets.
\item \textbf{Degenerate punishment}: contributes to $6$ of those $10$ A$\to$U downshifts, overlapping with mixed-output handling in $5$. In these high-$\alpha$ cells, the v1 keyword detector could count token-soup branches as successful A$\to$U flips when the soup happened to contain refusal vocabulary; v2 forces degenerate branches to flip false. Mistral fact A$\to$U $\alpha=40$ drops from $+34$pp (v1) to $+4$pp (v2) under $66\%$ branch degeneracy.
\item \textbf{Gate broadening} (code): the v2 invalidity-aware classifier accepts ``raises X'' baseline abstentions that the v1 keyword list missed. The clean-baseline gate enlarges, the same successful-flip count is divided by a larger denominator, and the gated rate dilutes. Mistral code U$\to$A: v1 $+71$pp on gateN $14$ becomes v2 $+44$pp on gateN $27$.
\item \textbf{Small-N reveal} ($3$ of $4$ up-shifts; reported as anecdotal): cells where v1 reported a positive number on a $2$--$5$ row gate that v2 either narrows (gate further drops to $1$--$4$) or widens slightly. These are not retained as evidence in either direction.
\item \textbf{Lexical FP catch in clean baseline}: not the dominant cause but contributes alongside mixed-output catch on math (small downshifts).
\end{itemize}

In the steering breadth sweep, $14$ of $48$ cells where the v1 keyword proxy reported positive hallucination reduction drop to non-positive under v2-deterministic. These cells are concentrated in small or fragile models (e.g.\ SmolLM2 fact, Mistral-Small fact) where lexical accidents on long-form generation inflated the legacy proxy.

\subsection{v2-only Diagnostics}
\label{app:protocol-audit-diagnostics}

The v2-deterministic JSONs include per-cell diagnostic fields not present in the legacy aggregation: \texttt{n\_mixed\_output\_overrides\_impos}, \texttt{n\_mixed\_output\_overrides\_rand}, \texttt{n\_degenerate\_impos}, \texttt{n\_degenerate\_rand}, and per-$\alpha$ \texttt{preservation\_failure} counts. \texttt{preservation\_failure} (an answerable-class branch that collapses into token-soup rather than preserving a normal answer) is folded into \texttt{wrong\_refusal\_rate\_A} for legacy metric compatibility, but exposed separately in the v2det diagnostic block. These diagnostics are how we identified the high-$\alpha$ collapse pattern across models and the family of mixed-output false positives that motivated the v2 refinement.

\subsection{Dose, Degeneration, and Tolerance-Window Diagnostics}
\label{app:protocol-audit-dose}

Tables~\ref{tab:dose-deg-utoa}--\ref{tab:dose-deg-atou} report every tested dose for the 16 structural cell-directions. Each entry is the invalidity-aware signal-minus-random gated effect $\Delta G$ in percentage points, followed by signal-branch and matched-random-branch degeneration rates in percent. Degenerate branches remain in the gate and count as non-flips, so collapse cannot inflate $\Delta G$.

\begin{table*}[t]
\centering
\scriptsize
\setlength{\tabcolsep}{3pt}
\begin{tabular}{@{}llrrrr@{}}
\toprule
Model & Dataset & $\alpha{=}5$ & $\alpha{=}10$ & $\alpha{=}20$ & $\alpha{=}40$ \\
\midrule
Mistral-7B & math800 & $27.6/2.0/2.0$ & $37.9/4.0/0.0$ & $20.7/30.0/4.0$ & $17.2/86.0/96.0$ \\
Mistral-7B & code800 & $7.4/0.0/0.0$ & $25.9/0.0/0.0$ & $44.4/0.0/0.0$ & $44.4/4.0/0.0$ \\
Gemma-3-4B & math800 & $28.6/8.0/2.0$ & $47.6/16.0/6.0$ & $33.3/40.0/86.0$ & $14.3/74.0/88.0$ \\
Gemma-3-4B & code800 & $6.7/2.0/2.0$ & $40.0/2.0/2.0$ & $0.0/4.1/2.0$ & $13.3/61.2/65.3$ \\
Qwen3-14B & math800 & $17.9/0.0/2.1$ & $3.6/12.5/2.1$ & $21.4/60.4/22.9$ & $3.6/97.9/100.0$ \\
Qwen3-14B & code800 & $52.2/12.5/2.1$ & $39.1/43.8/2.1$ & $13.0/72.9/22.9$ & $13.0/81.2/93.8$ \\
Qwen3-8B & math800 & $29.2/6.3/4.2$ & $29.2/6.3/6.3$ & $41.7/12.5/33.3$ & $-8.3/93.8/85.4$ \\
Qwen3-8B & code800 & $23.5/18.6/11.6$ & $17.6/34.9/14.0$ & $17.6/44.2/37.2$ & $-5.9/65.1/41.9$ \\
\bottomrule
\end{tabular}
\caption{Full U$\to$A structural dose and degeneration diagnostics. Entries are $\Delta G$ pp / signal degeneration \% / random degeneration \%.}
\label{tab:dose-deg-utoa}
\end{table*}

\begin{table*}[t]
\centering
\scriptsize
\setlength{\tabcolsep}{3pt}
\begin{tabular}{@{}llrrrr@{}}
\toprule
Model & Dataset & $\alpha{=}5$ & $\alpha{=}10$ & $\alpha{=}20$ & $\alpha{=}40$ \\
\midrule
Mistral-7B & math800 & $-2.0/0.0/0.0$ & $18.4/2.0/0.0$ & $32.7/6.0/12.0$ & $0.0/92.0/92.0$ \\
Mistral-7B & code800 & $2.1/0.0/0.0$ & $2.1/0.0/0.0$ & $12.5/2.1/0.0$ & $35.4/0.0/6.3$ \\
Gemma-3-4B & math800 & $2.1/6.1/10.2$ & $16.7/8.2/38.8$ & $0.0/83.7/28.6$ & $0.0/71.4/2.0$ \\
Gemma-3-4B & code800 & $2.1/4.2/0.0$ & $8.3/2.1/2.1$ & $35.4/0.0/16.7$ & $8.3/0.0/31.3$ \\
Qwen3-14B & math800 & $0.0/0.0/6.7$ & $7.0/6.7/11.1$ & $2.3/40.0/37.8$ & $0.0/82.2/86.7$ \\
Qwen3-14B & code800 & $4.3/0.0/2.1$ & $23.9/2.1/8.5$ & $37.0/8.5/46.8$ & $0.0/85.1/93.6$ \\
Qwen3-8B & math800 & $2.2/2.2/0.0$ & $6.7/8.9/0.0$ & $20.0/48.9/4.4$ & $0.0/100.0/100.0$ \\
Qwen3-8B & code800 & $0.0/0.0/0.0$ & $20.4/4.1/0.0$ & $22.4/20.4/4.1$ & $0.0/20.4/14.3$ \\
\bottomrule
\end{tabular}
\caption{Full A$\to$U structural dose and degeneration diagnostics. Entries are $\Delta G$ pp / signal degeneration \% / random degeneration \%.}
\label{tab:dose-deg-atou}
\end{table*}

All 10 anchor-quality directions reach $\Delta G\geq+30$pp while signal degeneration is at most $16\%$. Five of the six sub-threshold directions reach at least $40\%$ signal degeneration at a tested dose; Qwen3-8B code A$\to$U is the exception, with a $+22.4$pp peak and $20.4\%$ degeneration. Restricting each sub-threshold direction to tested doses below $25\%$ degeneration leaves a positive best effect in every case ($+6.7$ to $+23.5$pp). At $\alpha{=}20$, averaging the four structural directions within each model gives signal/random degeneration of $9.5/4.0\%$ for Mistral-7B, $31.9/33.3\%$ for Gemma-3-4B, $45.5/32.6\%$ for Qwen3-14B, and $31.5/19.8\%$ for Qwen3-8B. The matched-norm random branch therefore reproduces the Mistral-versus-rest tolerance gap, although the signal branch has additional direction-specific collapse in the weakest cells.

\begin{table*}[t]
\centering
\footnotesize
\setlength{\tabcolsep}{4pt}
\begin{tabular}{@{}llrrr@{}}
\toprule
Model & Dataset & $\cos(d_{\mathrm{imp}},d_{\mathrm{struct,behav}})$ & min. U$\to$A & min. A$\to$U \\
\midrule
Mistral-7B & math800 & 0.464 & 10 & 20 \\
Mistral-7B & code800 & 0.163 & 20 & 40 \\
Gemma-3-4B & math800 & 0.486 & 10 & $>40$ \\
Gemma-3-4B & code800 & 0.487 & 10 & 20 \\
Qwen3-14B & math800 & 0.214 & $>40$ & $>40$ \\
Qwen3-14B & code800 & 0.595 & 5 & 20 \\
Qwen3-8B & math800 & 0.326 & 20 & $>40$ \\
Qwen3-8B & code800 & 0.311 & $>40$ & $>40$ \\
\bottomrule
\end{tabular}
\caption{Behavior coupling and minimum tested dose reaching the $+30$pp anchor criterion. Treating the two never-reaching U$\to$A cells as tied above $40$, the descriptive Spearman correlation is $\rho=-0.86$ across eight structural cells. Among the four cells reaching the anchor in both directions, A$\to$U requires 2--4 times the U$\to$A dose.}
\label{tab:coupling-dose}
\end{table*}

The descriptive correlation between behavior coupling and minimum U$\to$A anchor dose is exploratory rather than significance-tested and does not predict best effect size. Recognition AUC also does not explain steering strength: Qwen3-14B and Qwen3-8B math have the two highest anchor AUCs (0.988 and 0.992) but are sub-threshold in at least one direction, whereas bidirectionally controllable Mistral-7B math has the lowest anchor AUC (0.909). With eight cells from four anchors across three model families, model tolerance, layer depth, and post-training design remain confounded; the diagnostics support a tolerance-window account but do not identify its cause.

\subsection{Anecdotal Cell Registry}
\label{app:protocol-audit-anecdotal}

Cells whose invalidity-aware gated denominator is $\leq 4$ in the 4-anchor intervention grid are reported as anecdotal and excluded from the headline tables in \S\ref{sec:findings-causal}. All sixteen anecdotal $\alpha$-rows fall in the U$\to$A direction on fact800; no math or code cell, and no A$\to$U fact cell, is anecdotal.

\begin{table*}[t]
\small
\centering
\begin{tabular}{lccrrrr}
\toprule
Anchor / dataset / layer & dir. & $\alpha$ & gateN & sig\% & rnd\% & note \\
\midrule
mistral / fact800 / L17       & U$\to$A & 5  & 0 & N/A   & N/A   & empty gate \\
mistral / fact800 / L17       & U$\to$A & 10 & 0 & N/A   & N/A   & empty gate \\
mistral / fact800 / L17       & U$\to$A & 20 & 0 & N/A   & N/A   & empty gate \\
mistral / fact800 / L17       & U$\to$A & 40 & 0 & N/A   & N/A   & empty gate (deg.\ 30/44) \\
gemma3\_4b / fact800 / L16    & U$\to$A & 5  & 1 & 0     & 0     & single-row gate \\
gemma3\_4b / fact800 / L16    & U$\to$A & 10 & 1 & 100   & 0     & misleading n=1 \\
gemma3\_4b / fact800 / L16    & U$\to$A & 20 & 1 & 100   & 100   & single-row gate \\
gemma3\_4b / fact800 / L16    & U$\to$A & 40 & 1 & 100   & 0     & misleading n=1 \\
qwen3\_14b / fact800 / L25    & U$\to$A & 5  & 4 & 50    & 0     & 4-row gate \\
qwen3\_14b / fact800 / L25    & U$\to$A & 10 & 4 & 100   & 0     & misleading n=4 \\
qwen3\_14b / fact800 / L25    & U$\to$A & 20 & 4 & 75    & 25    & 4-row gate \\
qwen3\_14b / fact800 / L25    & U$\to$A & 40 & 4 & 75    & 0     & deg-rnd 96\% \\
qwen3\_8b / fact800 / L21     & U$\to$A & 5  & 2 & 50    & 0     & 2-row gate \\
qwen3\_8b / fact800 / L21     & U$\to$A & 10 & 2 & 50    & 50    & 2-row gate \\
qwen3\_8b / fact800 / L21     & U$\to$A & 20 & 2 & 0     & 50    & sign flip is noise \\
qwen3\_8b / fact800 / L21     & U$\to$A & 40 & 2 & 50    & 100   & 2-row gate \\
\bottomrule
\end{tabular}
\caption{Anecdotal cells (gated denominator $\leq 4$) in the 4-anchor intervention grid. All sixteen rows are in the fact U$\to$A direction. The pattern is uniform across anchors: SQuAD-style passage-grounded epistemic unanswerability does not produce many ``clean abstention without appended concrete answer'' baselines under the invalidity-aware rubric, leaving the U$\to$A direction structurally unmeasurable on this domain. v1's keyword detector accepted lexical noise (``no'', ``not'', ``unknown'' fragments embedded in long-form passage continuations) and reported a positive U$\to$A fact $\Delta$G; v2 surfaces that the invalidity-aware-clean baseline population is structurally too small to support a per-cell $\Delta$G measurement on this direction.}
\label{tab:anecdotal-fact}
\end{table*}

\subsection{Steering Breadth Per-Domain Numbers}
\label{app:protocol-audit-breadth}

Re-aggregating the 48-cell steering breadth sweep (16 models $\times$ 3 datasets) under the v2-deterministic invalidity-aware regex proxy (\S\ref{sec:findings-causal}), mean hallucination-rate reduction on U at best $\alpha$ (relative to the $\alpha{=}0$ baseline) is $+0.13$ on math800 ($15/16$ models positive), $+0.09$ on code800 ($12/16$), and $+0.004$ on fact800 ($4/16$): a math $>$ code $>$ fact gradient. This metric is hallucination reduction, not a signal-minus-random gain, so it is not a comparable effect size to the intervention gated $\Delta$G and ranks math vs.\ code differently from the intervention grid; the two nonetheless converge on fact as the weakest domain.

\subsection{Framing}

For the intervention grid, v2 uses candidate labels from this mixed workflow and applies provisional second-pass audit fills in nine cells; the three Qwen3-8B cells use candidate-label passthrough. For the steering breadth sweep, v2 is a deterministic re-aggregation proxy aligned to the intervention rubric (no relabeling on either v1 or v2). The trade is to lose some peak effect-size magnitude (math/code intervention $\Delta$G in the $+33$ to $+52$pp range under v2 vs.\ $+42$ to $+100$pp under v1) in exchange for two methodological gains: (i) cross-protocol convergence on fact-as-boundary that v1 single-protocol headline could not surface, and (ii) explicit refusal-only universality measurement that directly addresses the alternative reading of $d_{\mathrm{imp}}$ as a generic refusal-vocabulary axis. The substantive conclusions of \S\ref{sec:findings-causal} (math/code admit causal control, with Mistral-7B as the bidirectional keystone; fact is a structural boundary; recognition is distinct from refusal in both geometry and behavior) hold under both versions; v2 narrows the claim where v1 was vulnerable to keyword artifacts and gates the fact-domain claim where v1 reported a small-$N$ anecdote.

\section{Behavior-Defined Invalidity-Aware Direction}
\label{app:struct-behav}

This appendix reports the per-cell numbers behind the behavior-defined direct comparison of \S\ref{sec:findings-ortho}. The direction $d_{\mathrm{struct,behav}}$ is a mean-difference of prompt last-token hidden states on U-class (impossible) prompts at the clean baseline, contrasting generations the model's output labels \emph{invalidity-aware} (acknowledges the problem has no valid value, is undefined, or cannot be computed) against generations that answer anyway. It is computed in the same A-class PCA null subspace, with the same $V_A$, as the matched-layer $d_{\mathrm{imp}}$ for that cell. We report it on the four-anchor intervention grid $\times$ \{math800, code800\} (eight cells). The six-cell primary non-Qwen3-8B subset uses candidate labels from the mixed workflow plus provisional second-pass audit fills; the two Qwen3-8B cells use candidate labels without second-pass overrides.

\begin{table*}[t]
\centering
\small
\setlength{\tabcolsep}{4pt}
\resizebox{\textwidth}{!}{%
\begin{tabular}{@{}llccllllcc@{}}
\toprule
Model & Dataset & $L$ & $n_y$/$n_n$ & label & $\cos(d_{\mathrm{i}},d_{\mathrm{sb}})$ [95\% CI] & $\cos(d_{\mathrm{i}},d_{\mathrm{rs}})$ & $\cos(d_{\mathrm{sb}},d_{\mathrm{rs}})$ [95\% CI] & LOO & expl. \\
\midrule
Mistral-7B & math800 & 15 & 29/21 & pass-2 & $+0.464$ $[+0.196, +0.522]$ & $+0.130$ & $+0.079$ $[+0.023, +0.097]$ & 0.80 & --- \\
Mistral-7B & code800 & 15 & 27/23 & pass-2 & $+0.163$ $[-0.042, +0.315]$ & $+0.098$ & $+0.013$ $[-0.035, +0.054]$ & 0.66 & \textbf{yes} \\
Gemma-3-4B & math800 & 16 & 21/29 & pass-2 & $+0.486$ $[+0.270, +0.524]$ & $+0.153$ & $+0.124$ $[+0.049, +0.149]$ & 0.84 & --- \\
Gemma-3-4B & code800 & 15 & 15/34 & pass-2 & $+0.487$ $[+0.264, +0.529]$ & $+0.152$ & $+0.090$ $[+0.038, +0.102]$ & 0.91 & --- \\
Qwen3-14B & math800 & 25 & 28/20 & pass-2 & $+0.214$ $[-0.009, +0.378]$ & $+0.091$ & $+0.060$ $[+0.009, +0.090]$ & 0.84 & --- \\
Qwen3-14B & code800 & 24 & 23/25 & pass-2 & $+0.595$ $[+0.385, +0.675]$ & $+0.139$ & $+0.087$ $[+0.048, +0.104]$ & 0.91 & --- \\
Qwen3-8B & math800 & 21 & 24/24 & cand. & $+0.326$ $[+0.060, +0.459]$ & $+0.113$ & $+0.049$ $[+0.003, +0.073]$ & 0.78 & --- \\
Qwen3-8B & code800 & 19 & 17/26 & cand. & $+0.311$ $[-0.013, +0.467]$ & $+0.121$ & $+0.116$ $[+0.050, +0.136]$ & 0.73 & --- \\
\midrule
\multicolumn{5}{l}{\textbf{Mean (8 cells)}} & $+0.381$ & $+0.125$ & $+0.077$ & 0.81 & \\
\bottomrule
\end{tabular}%
}
\caption{Per-cell A-null primary subspace cosines for $d_{\mathrm{struct,behav}}$ (directly comparable to \S\ref{sec:findings-ortho}). Abbreviations: $d_{\mathrm{i}}\equiv d_{\mathrm{imp}}$, $d_{\mathrm{sb}}\equiv d_{\mathrm{struct,behav}}$, $d_{\mathrm{rs}}\equiv d_{\mathrm{ref,safety}}$. CIs are 1000-iteration bootstrap over the yes/no label set only, with $d_{\mathrm{i}}$ and $d_{\mathrm{rs}}$ held fixed; $\cos(d_{\mathrm{i}},d_{\mathrm{rs}})$ is therefore a point estimate. LOO AUC is the held-out separability metric $\max(\mathrm{AUC},1-\mathrm{AUC})$. A cell is flagged exploratory (expl.) when LOO AUC $<0.65$ or bootstrap self-cosine median $<0.80$. Six-cell primary means (excluding the two Qwen3-8B candidate-only cells) are $0.401$, $0.127$, and $0.076$ for the three cosines. ``pass-2'' denotes provisional LLM-assisted second-pass audit fills, not completed human adjudication.}
\label{tab:struct-behav-anull}
\end{table*}

\begin{table*}[t]
\centering
\small
\setlength{\tabcolsep}{4pt}
\resizebox{\textwidth}{!}{%
\begin{tabular}{@{}lllllcc@{}}
\toprule
Model & Dataset & $\cos(d_{\mathrm{i}},d_{\mathrm{sb}})$ [95\% CI] & $\cos(d_{\mathrm{i}},d_{\mathrm{rs}})$ & $\cos(d_{\mathrm{sb}},d_{\mathrm{rs}})$ [95\% CI] & LOO & expl. \\
\midrule
Mistral-7B & math800 & $+0.562$ $[+0.350, +0.584]$ & $+0.272$ & $+0.181$ $[+0.091, +0.207]$ & 0.82 & --- \\
Mistral-7B & code800 & $+0.320$ $[+0.036, +0.430]$ & $+0.245$ & $+0.115$ $[-0.011, +0.178]$ & 0.62 & \textbf{yes} \\
Gemma-3-4B & math800 & $-0.229$ $[-0.503, +0.602]$ & $+0.531$ & $-0.244$ $[-0.404, +0.333]$ & 0.73 & --- \\
Gemma-3-4B & code800 & $+0.932$ $[+0.772, +0.937]$ & $+0.781$ & $+0.780$ $[+0.631, +0.791]$ & 0.78 & --- \\
Qwen3-14B & math800 & $+0.280$ $[+0.074, +0.458]$ & $+0.251$ & $+0.138$ $[+0.059, +0.193]$ & 0.86 & --- \\
Qwen3-14B & code800 & $+0.590$ $[+0.356, +0.652]$ & $+0.234$ & $+0.152$ $[+0.089, +0.172]$ & 0.87 & --- \\
Qwen3-8B & math800 & $+0.509$ $[+0.240, +0.584]$ & $+0.238$ & $+0.158$ $[+0.086, +0.173]$ & 0.80 & --- \\
Qwen3-8B & code800 & $+0.373$ $[+0.072, +0.472]$ & $+0.278$ & $+0.157$ $[+0.058, +0.183]$ & 0.72 & --- \\
\midrule
\multicolumn{2}{l}{\textbf{Mean (8 cells)}} & $+0.417$ & $+0.354$ & $+0.180$ & 0.78 & \\
\bottomrule
\end{tabular}%
}
\caption{Full-space robustness pass: no A-null projection on $d_{\mathrm{i}}$ or $d_{\mathrm{sb}}$; $d_{\mathrm{rs}}$ in its native full space. Abbreviations and CI convention as in Table~\ref{tab:struct-behav-anull}. The pass agrees with A-null in aggregate (eight-cell mean $0.417$; six-cell primary mean $0.409$) except Gemma-3-4B / math800, whose full-space $d_{\mathrm{i}}$ is contaminated by task-related variance (cosine $-0.23$ with a sign-spanning CI); its A-null value is a clean $+0.49$. This is the failure mode the A-null subspace was designed to remove, and the reason the primary measurement is in A-null.}
\label{tab:struct-behav-full}
\end{table*}

\textbf{Construction and uncertainty.} $d_{\mathrm{struct,behav}}$ is built from prompt last-token hidden states (the representation in which $d_{\mathrm{imp}}$ and $d_{\mathrm{ref,safety}}$ are also defined), so the comparison is between pre-generation prompt-time directions, not generation-time circuits. Table~\ref{tab:struct-behav-anull} projects $d_{\mathrm{imp}}$ and $d_{\mathrm{struct,behav}}$ into the same A-class PCA null subspace $V_A$ used for that cell's $d_{\mathrm{imp}}$; Table~\ref{tab:struct-behav-full} applies no null projection. Bootstrap 95\% CIs resample the yes/no invalidity-aware label set only (1000 iterations), holding $d_{\mathrm{imp}}$ and $d_{\mathrm{ref,safety}}$ fixed, so $\cos(d_{\mathrm{imp}}, d_{\mathrm{ref,safety}})$ is a point estimate here (its main-grid bootstrap CI is the one reported in \S\ref{sec:findings-ortho}). LOO AUC is reported as $\max(\mathrm{AUC}, 1-\mathrm{AUC})$ on held-out yes/no samples and is a \emph{separability} metric: it measures whether the direction distinguishes invalidity-aware from not; it does not check the yes$-$no orientation, which fixes the sign by construction.

\textbf{Strict refusal-only is absent.} Under a strict refusal-only criterion (explicit ``I cannot answer'' / ``this is unanswerable'' phrasing), 0 of 50 U-class prompts qualify in every one of the eight cells, so a natural strict-refusal direction cannot be constructed in this domain. This sparsity is the supporting evidence for using the broader invalidity-aware criterion in \S\ref{sec:findings-ortho}, and is itself consistent with the paper's claim that the model rarely refuses structurally impossible prompts even when it internally recognizes them.

\textbf{Reproducibility note.} A-null cosines carry a $\sim$0.005 noise band because sklearn's default PCA solver is non-deterministic at this matrix shape ($n_{\mathrm{samples}}{=}400$, $n_{\mathrm{features}}{\in}[2560,5120]$, $k{=}100$); full-space cosines (no PCA) reproduce the existing \texttt{direction\_comparison} values exactly ($\Delta=0$). The band is an order of magnitude below the headline gap ($0.40$ vs.\ $0.13$); pinning \texttt{svd\_solver="full"} with a fixed \texttt{random\_state} makes future reproductions exact.

\textbf{Base-model recognition (\S\ref{sec:findings-pretrain}).} Table~\ref{tab:base-imp-auc} reports the one-dimensional A-null $d_{\mathrm{imp}}$ detection AUC on base-model math800 at each base checkpoint's matched layer, for the six base/instruct pairs of \S\ref{sec:findings-pretrain}. The mean is $0.977$, comparable to the instruct main grid (\S\ref{sec:findings-detection}): impossibility is already linearly recoverable from a single direction before instruction tuning.

\begin{table}[h]
\centering
\small
\setlength{\tabcolsep}{6pt}
\begin{tabular}{@{}lcc@{}}
\toprule
Base checkpoint & matched $L$ & $d_{\mathrm{imp}}$ AUC \\
\midrule
Qwen2.5-7B-Base    & 18 & 0.944 \\
Qwen2.5-14B-Base   & 34 & 0.982 \\
Qwen2.5-32B-Base   & 53 & 0.985 \\
Qwen3-8B-Base      & 21 & 0.983 \\
Qwen3-14B-Base     & 25 & 0.989 \\
Llama-3.1-70B-Base & 31 & 0.982 \\
\midrule
\textbf{Mean}      &    & \textbf{0.977} \\
\bottomrule
\end{tabular}
\caption{Base-model $d_{\mathrm{imp}}$ detection AUC on math800 (A-null MeanDiff, one direction) at each base checkpoint's matched layer, for the six base/instruct pairs of \S\ref{sec:findings-pretrain}. Values are the \texttt{auc\_impossibility\_on\_dataset} field of the per-model \texttt{direction\_comparison} records; mean $0.977$.}
\label{tab:base-imp-auc}
\end{table}

\textbf{Base/instruct pair detail (\S\ref{sec:findings-pretrain}).} The six pairs are Qwen2.5-7B/14B/32B-Instruct, Qwen3-8B, and Qwen3-14B vs.\ their bases, plus Llama-3.3-70B-Instruct vs.\ Llama-3.1-70B-Base. The first five verify on both sides; the Llama-70B base uses a proxy $d_{\mathrm{ref,safety}}$ (\texttt{base\_behavior\_verified}{=}\texttt{False}) because Llama-3.1-70B-Base rarely refuses harmful prompts, though Meta confirms Llama-3.3-70B-Instruct shares the Llama-3.1-70B pretraining checkpoint, making it the grid's only vendor-confirmed post-training-only contrast. Qwen3-32B has no public base release as of May 2026, so the 32B comparison substitutes Qwen2.5-32B (which behavior-verifies); this scope split is deliberate. Within-family $\cos(d_{\mathrm{imp}}, d_{\mathrm{ref,safety}})$ values: Qwen2.5-7B base 0.122 / instruct 0.114; Qwen2.5-14B 0.056 / 0.085; Qwen2.5-32B 0.066 / 0.071.

\section{Multi-Dimensional Geometry Robustness}
\label{app:multidim}

The main analysis summarizes each representation with one direction and cosine similarity. We therefore computed a direct linear-subspace check from saved representations on Mistral-7B and Qwen3-14B, each on math800 and code800. For each cell, the $k$-dimensional impossibility subspace contains the top-$k$ singular directions of U-class deviations from the A-class mean in A-null space. Its first component is fit on the training split and evaluated on held-out A/U states. We compare it with the behavior-verified safety-refusal direction and with a $k$-dimensional refusal subspace constructed analogously from harmful-prompt deviations from the harmless mean.

\begin{table*}[t]
\centering
\scriptsize
\setlength{\tabcolsep}{3pt}
\resizebox{\textwidth}{!}{%
\begin{tabular}{@{}llcccccc@{}}
\toprule
Model & Dataset & $E(d_{\mathrm{imp}})$ & first-comp. AUC & $E(d_{\mathrm{ref}})$ full (\%) & $E(d_{\mathrm{ref}})$ A-null (\%) & min. angle (deg.) & random first $\cos$ \\
 & & $k{=}5/10$ & & $k{=}5/10$ & $k{=}5/10$ & $k{=}5/10$ & $k{=}5/10$ \\
\midrule
Mistral-7B & math800 & $0.889/0.956$ & 0.740 & $1.94/2.41$ & $2.31/2.86$ & $80.3/77.2$ & $0.058/0.088$ \\
Mistral-7B & code800 & $0.909/0.942$ & 0.816 & $1.54/2.00$ & $1.89/2.45$ & $82.0/79.4$ & $0.057/0.089$ \\
Qwen3-14B & math800 & $0.969/0.992$ & 0.833 & $1.23/2.14$ & $1.41/2.44$ & $81.7/79.0$ & $0.050/0.079$ \\
Qwen3-14B & code800 & $0.991/0.992$ & 0.929 & $2.09/3.41$ & $2.37/3.87$ & $78.0/74.1$ & $0.051/0.079$ \\
\bottomrule
\end{tabular}%
}
\caption{Multi-dimensional robustness at $k\in\{5,10\}$. $E(d_{\mathrm{imp}})$ is the fraction of the one-dimensional impossibility direction's energy captured by the impossibility subspace. $E(d_{\mathrm{ref}})$ is refusal-direction projector energy before and after projecting and renormalizing refusal in the same A-null space. ``min. angle'' compares the impossibility and behavior-verified refusal subspaces; the final column is the equal-dimensional random-subspace mean for the first principal cosine.}
\label{tab:multidim}
\end{table*}

The impossibility subspace captures $0.89$--$0.99$ of $d_{\mathrm{imp}}$'s energy, and its first component remains discriminative on held-out data in all four cells (AUC $0.74$--$0.93$). Projecting $d_{\mathrm{ref,safety}}$ onto this subspace captures only $1.2$--$2.1\%$ of its energy at $k{=}5$ and $2.0$--$3.4\%$ at $k{=}10$, compared with random-direction means of $0.1$--$0.25\%$. Projecting and renormalizing refusal in the same A-null space yields $1.4$--$2.4\%$ and $2.4$--$3.9\%$, respectively. Thus at least $96\%$ of the refusal direction lies outside even the 10-dimensional impossibility subspace. The smallest behavior-verified principal angle is at least $74^\circ$ in every cell (first principal cosine $0.14$--$0.27$, versus random means $0.05$--$0.09$). Because the angle check compares an A-null impossibility subspace with a full-space refusal subspace, the projector-energy measurements are the primary same-space evidence. The overlap is small but above chance: on these four cells, the low-overlap conclusion survives a 5--10-dimensional linear-subspace analysis, while nonlinear or more distributed abstention representations remain possible.

\section{Robustness Controls (Ruling Out Alternatives)}
\label{app:robustness}

This appendix gives the full per-control evidence summarized in \S\ref{sec:char}.

\textit{Not length.} U-class prompts can be shorter or simpler than their A counterparts, and on natural-distribution data the concern sharpens (length alone reaches near-ceiling AUC on AbstentionBench-GSM8K). Restricting that benchmark to a length-matched subset, where a length-only classifier is neutralized to AUC 0.500, leaves CosNSRT detection at 0.799, comparable to the full-set value. \textit{Not difficulty.} A difficulty axis correlated with answerability could fake both \S\ref{sec:findings-ortho} and \S\ref{sec:findings-causal}. A difficulty-controlled split within the answerable class gives $d_{\mathrm{imp}}$ AUC 0.61 as a hard-vs-easy classifier vs.\ 0.96 on impossibility ($\Delta = +0.35$); the direction is far more sensitive to answerability than to difficulty.

\textit{Not a category-specific artifact.} The 22-cell global AUCs average over 16 math800 + 8 code800 categories, and could be propped up by a few easy categories. Within-cell Spearman $\rho$ between NS\_SNR (a label-free null-space class-separation statistic) and per-category CosNSRT AUC has mean 0.730 across the 22 cells, with 18/22 cells reaching $p < 0.05$; the global mean is an interpretable summary, not a few-easy-cells artifact.

\textit{Not only one model scale.} The recognition signal and the orthogonality could be small-model quirks. Across the 22 cells, detection AUC ranges 0.841--0.993 and the orthogonality regime tightens at the largest scales: 24B Mistral-Small on math800 has cos = 0.065 with bootstrap CI $[0.059, 0.071]$, and the 32B / 70B math800 cosines remain in-band at $\approx 0.098$. Both regularities strengthen with scale rather than washing out.

\textit{Not a generic unanswerability axis.} The within-minus-cross drop is mean 0.080 (range $[-0.005, 0.207]$) across 22 directional drops, arguing against a single generic unanswerability axis. In an auxiliary earlier-grid natural-transfer check (8 models; not 11-model headline evidence), math800-fit $d_{\mathrm{imp}}$ transfers partially under dot-product NSRT scoring to AbstentionBench-GSM8K (0.64--0.98) and FalseQA (0.59--0.90). Separately, mean CosNSRT AUC remains 0.799 after length matching AbstentionBench-GSM8K, where the length-only AUC is 0.500. These are detection-only boundary tests, not evidence that steering transfers. We therefore treat structural impossibility as the clean setting; fact800 and FalseQA are boundary cases, not equivalent evidence.

\end{document}